\documentclass[10pt,twocolumn,letterpaper]{article}

\usepackage{wacv}              

\usepackage[accsupp]{axessibility}  

\usepackage{multirow}
\usepackage{xcolor}
\usepackage{arydshln}
\usepackage{makecell}
\definecolor{llamaindigo}{RGB}{99, 102, 241}   
\definecolor{qwenamber}{RGB}{245, 158, 11}     
\definecolor{robobrain}{RGB}{148, 163, 184}    

\usepackage{colortbl}
\colorlet{trajperf}{blue!8}
\colorlet{trajeff}{orange!10}
\colorlet{affperf}{teal!8}
\colorlet{affeff}{orange!10}
\definecolor{ourgreen}{RGB}{168, 208, 168}  

\renewcommand{\ie}{\emph{i.e.},}

\definecolor{wacvblue}{rgb}{0.21,0.49,0.74}
\usepackage[pagebackref,breaklinks,colorlinks,allcolors=wacvblue]{hyperref}

\def\wacvPaperID{393} 
\def\confName{WACV}
\def\confYear{2027}

\title{Decoding Geometry, Not Tokens: Efficient VLMs for Robot Affordance and Trajectory Prediction}

\author{
Akila Ayanthi$^{1,2}$ \quad
Darshana Priyasad$^{1}$ \quad
Tharindu Fernando$^{1}$ \quad
Dimity Miller$^{1,3}$ \\
Sridha Sridharan$^{1}$ \quad
Clinton Fookes$^{1}$ \quad
Peyman Moghadam$^{2}$ \\[6pt]
$^{1}$SAIVT, Queensland University of Technology, Australia \quad
$^{2}$CSIRO Robotics, Australia \\
$^{3}$QUT Centre for Robotics, Australia \\[4pt]
{\tt\small \{a.mudiyanselage, dp.don, t.warnakulasuriya, d24.miller, s.sridharan, c.fookes\}@qut.edu.au} \\
{\tt\small peyman.moghadam@csiro.au}
}

\begin{document}
\maketitle
\begin{abstract}
Predicting where to grasp and how to move is fundamentally a geometry problem. Vision-language models treat it as language generation instead. They tokenize spatial coordinates into text and decode them autoregressively. This mechanism ignores spatial structure and truncates continuous precision. Errors compound across coordinate tokens during decoding. In this paper, we show that compact off-the-shelf VLMs with 0.5--2B parameters already encode rich spatial information in their intermediate hidden states. Autoregressive decoding is the dominant bottleneck for trajectory prediction, while affordance prediction depends additionally on feature depth and decoder architecture. We expose the latent spatial information with a lightweight, geometry-compatible prediction head that attaches to a frozen backbone. It reads features from an intermediate hidden layer, conditions a learned query on the task instruction and refines it via cross-attention over the visual tokens to regress affordance boxes and trajectory waypoints directly as continuous values, without modifying the backbone. We evaluate on ShareRobot-Bench and AGD20K. Our 2B variant reduces trajectory error by over 60\% and improves ShareRobot-Bench affordance mAP by 85\% relative to the autoregressive baseline. It approaches RoboBrain on trajectory prediction (0.251 vs 0.191 DFD) at 3.8$\times$ higher throughput. These results demonstrate a favorable accuracy--efficiency trade-off for instruction-conditioned 2D geometric prediction.
\end{abstract}
    
\section{Introduction}

\begin{figure}
    \centering
    \includegraphics[width=0.8\linewidth]{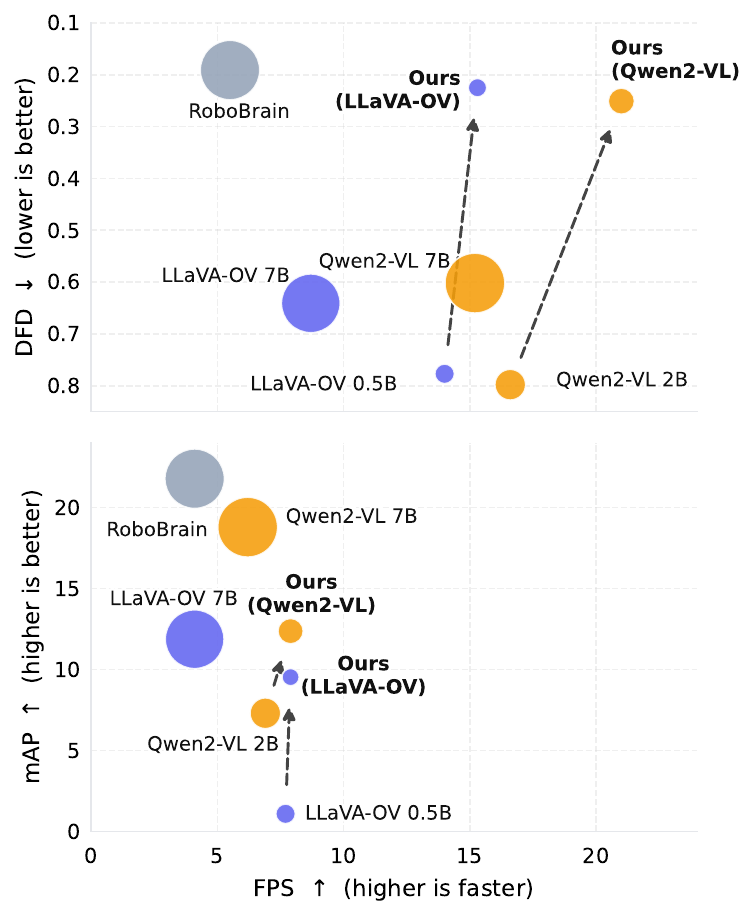}
    \vspace{-2mm}
    \caption{\textbf{Accuracy, efficiency, and model size.} Each bubble 
represents a model; bubble size encodes the number of active parameters. 
Color denotes model family: \textcolor{llamaindigo}{LLaVA-OV}, 
\textcolor{qwenamber}{Qwen2-VL}, \textcolor{robobrain}{RoboBrain}. 
Dashed arrows indicate improvement over the corresponding base VLM. 
\textbf{(Top)} Trajectory prediction DFD ($\downarrow$) vs.\ inference speed (FPS on H100) on ShareRobot-Bench-Trajectory. 
\textbf{(Bottom)} Affordance mAP ($\uparrow$) vs.\ inference speed (FPS on H100) on AGD20K. 
Our method outperforms both base VLMs, approaches larger-model accuracy, and retains favorable parameter and inference efficiency.}
    \label{fig:fps_params}
    \vspace{-2mm}
\end{figure}

Vision-language models (VLMs) are powerful tools for general-purpose reasoning, demonstrating strong performance across visual question answering \cite{chen2024internvl,li2024llava}, image captioning \cite{li2023blip,wang2024qwen2}, and scene understanding \cite{gholami2025spatial, selvam2025simcache}. 

However, using VLMs efficiently for robotic manipulation remains challenging, particularly for instruction-conditioned geometric prediction prior to physical execution. Manipulation requires spatially precise outputs describing \emph{where} to interact with the scene (\ie{} affordance prediction) and \emph{how} to move there (\ie{} trajectory prediction), whereas standard VLMs are designed primarily for tokenized language generation. Practical robotic systems also operate under constrained compute budgets \cite{grover2026embodied, zhou2026characterizing}, while modern VLMs can contain billions of parameters \cite{alayrac2022flamingo, bai2025qwen25vltechnicalreport}. We therefore ask: \emph{can compact VLMs provide accurate geometric predictions for robotic affordance and trajectory prediction, while retaining efficient inference?}


%
Existing approaches typically use VLMs for robotic manipulation by producing tokenized integers that are decoded autoregressively~\cite{qian2024affordancellm,yuan2024robopoint,ji2025robobrain}. This effectively treats geometric prediction as a language generation problem. Although this allows direct utilization of pre-trained VLMs and their existing architectures, it introduces a structural mismatch between the prediction mechanism and the task objective: Token-level cross-entropy learning objectives do not penalize geometric distance, finite token vocabularies do not allow for continuous spatial outputs and introduce quantization error, and autoregressive generation compounds errors across coordinate tokens. While larger VLM capacity, particularly when paired with large-scale training on robotics data~\cite{ji2025robobrain, driess2026knowledge, hancock2025actions} can yield performance gains for robotics manipulation in spite of this mismatch, this is at the cost of substantially lower inference speeds and higher compute requirements.

In this paper, we investigate how compact, off-the-shelf VLMs can be repurposed for geometric prediction. We replace token-based coordinate generation with a lightweight geometry-compatible decoder that directly regresses affordance boxes and trajectory waypoints from a \emph{frozen} VLM. Our systematic ablations show that direct continuous prediction is the primary source of improvement for trajectory prediction, while affordance prediction depends more strongly on the interaction between prediction interface, representation depth, and decoder architecture. Layer-wise analysis further reveals that useful spatial information is already present across the VLM representation hierarchy, although its accessibility varies with task and depth. Our contribution is therefore not the DETR architecture itself, but identifying the decoding-interface mismatch and showing how existing geometric representations can be exploited efficiently.

Overall, we make three empirically validated claims:
\begin{itemize}
    \item A geometry-compatible prediction interface substantially improves affordance and trajectory prediction while preserving the efficiency of compact VLMs (Sec.~\ref{sec:main_results}).

    \item Direct continuous prediction is the primary source of improvement for trajectory prediction, while affordance prediction depends on the interaction between prediction interface, representation depth, and decoder architecture (Sec.~\ref{sec:disentangle_abl}).


    \item Compact VLMs already encode useful spatial information across their representation hierarchy, with its accessibility varying across tasks and model depth (Sec.~\ref{sec:spatial_information}).
\end{itemize}

\section{Related Work}
\label{sec:related_work}

\subsection{Vision-Language Models}

Vision-language models (VLMs) pretrained on internet-scale datasets have demonstrated remarkable capabilities across tasks, including visual question answering \cite{chen2024internvl,li2024llava} and object localization \cite{doveh2025teaching,schaumloffel2026mechanisms}. Scaling to tens of billions of parameters, for example, Flamingo-80B \cite{alayrac2022flamingo}, InternVL3-78B \cite{zhu2025internvl3}, Qwen2.5-VL-72B \cite{bai2025qwen25vltechnicalreport}, has driven strong zero and few‑shot performance, but at high computational cost for deployment. 


Recent efforts have focused on improving efficiency through architectural design and model compression, leading to compact models such as LLaVA-OneVision (0.5B, 7B) \cite{li2024llava-ov}, Qwen2-VL (2B, 4B, 7B) \cite{wang2024qwen2}, Molmo (1B, 7B) \cite{deitke2025molmo}. Despite their strong semantic capabilities, most VLMs are trained using autoregressive next-token prediction and are therefore optimized for language generation rather than geometric reasoning. Consequently, spatially demanding tasks often require full fine-tuning or parameter-efficient adaptation to learn task-specific geometric representations \cite{ma2024spatialpin, chen2024spatialvlm}.
In this work, we instead treat compact VLMs as frozen feature extractors and ask whether their internal multimodal representations already contain sufficient geometric information for affordance and trajectory prediction. We use lightweight, geometry‑aware decoders on top of frozen features, avoiding any modifications to the backbone.

\subsection{VLMs for Embodied Tasks}

VLMs have been widely adopted in robotics for representation learning \cite{nair2022r3m,karamcheti2023language,huang2024manipvqa}, high-level planning \cite{zawalski2024robotic,sun2025emma}, and reward or feedback generation \cite{sontakke2023roboclip,zhang2023grounding}. More recently, Vision-Language-Action (VLA) models have extended this paradigm by directly mapping visual and language observations to low-level actions \cite{team2024octo,o2024open, lee2025molmoact}. Closely related embodied foundation models further integrate perception, reasoning, memory, and planning capabilities \cite{ji2025robobrain,tan2025roboos,yuan2024robopoint}. Examples include AffordanceLLM \cite{qian2024affordancellm}, which predicts affordance tokens for dense mask generation, and models such as LLARVA \cite{niu2024llarva}, RoboBrain \cite{ji2025robobrain}, and RoboPoint \cite{yuan2024robopoint}, which rely on extensive backbone adaptation using large embodied datasets.

Although these approaches achieve impressive generalization, they inherit the computational cost of large VLM backbones and remain challenging to deploy on resource-constrained platforms \cite{zhou2026characterizing,zhong2025survey}. Recent studies identify the VLM backbone itself as a major computational bottleneck \cite{grover2026embodied, yu2025survey} and motivate efficient variants such as EdgeVLA \cite{budzianowski2025edgevla, shukor2025smolvla}.
Unlike prior work that designs new VLA architectures or heavily fine‑tunes backbones, we use compact off‑the‑shelf VLMs as frozen perception modules and learn lightweight geometry‑aware decoders for affordance and trajectory prediction.
\section{Methodology}
\label{sec:method}

\begin{figure*}
    \centering
    \includegraphics[width=0.8\linewidth]{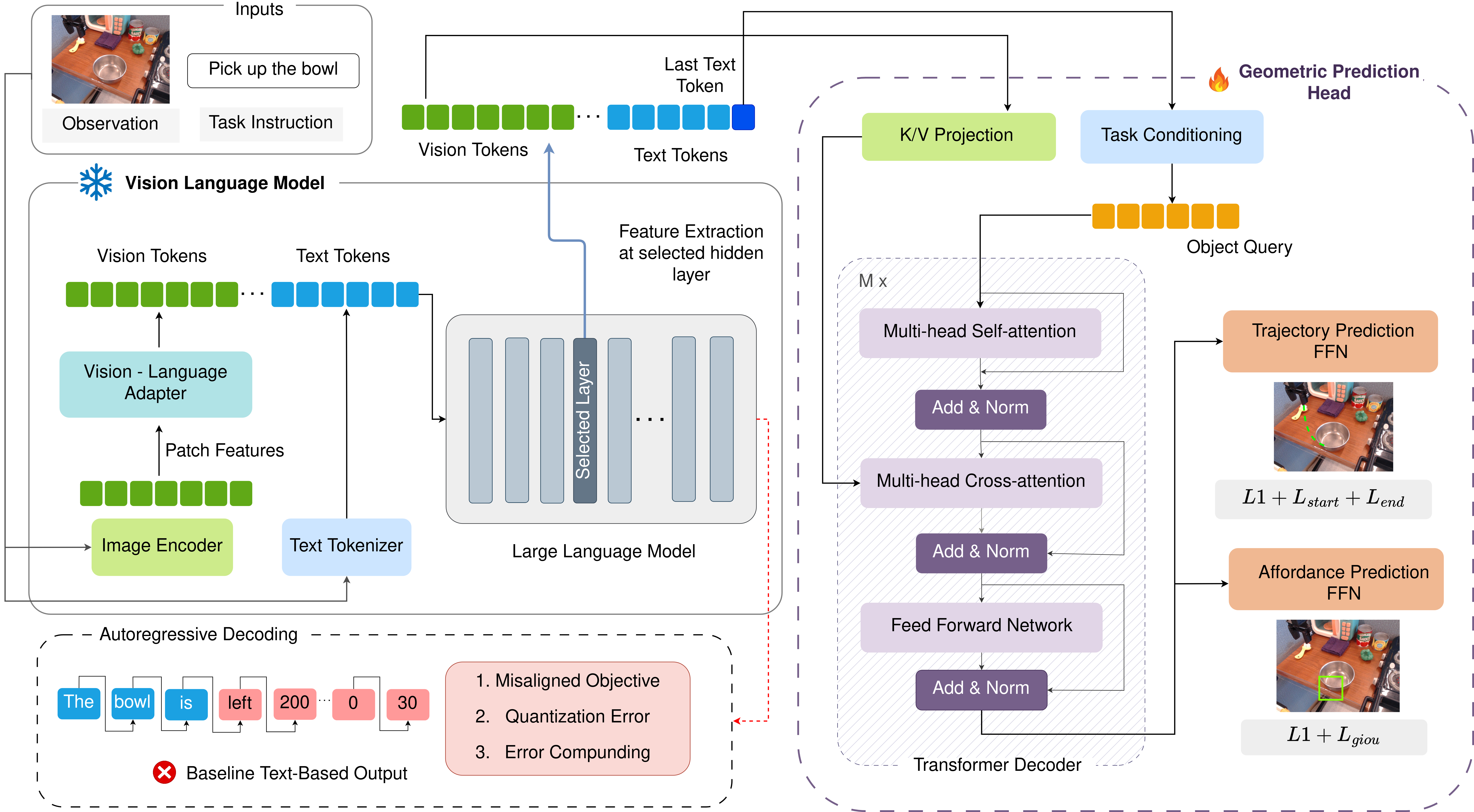}
    \caption{\textbf{Architecture overview.} A frozen VLM processes an RGB image and task instruction; hidden states from an intermediate layer are passed to a lightweight geometric prediction head. A DETR-style transformer decoder, conditioned on the instruction via the final text token hidden state, refines a learned query against the visual memory tokens. Two independent FFNs decode the query into continuous outputs: affordance boxes (L1+GIoU loss) and trajectory waypoints (L1+startpoint+endpoint loss). No gradient flows into the VLM.}
    \label{fig:overview}
    \vspace{-2mm}
\end{figure*}

Our goal is to bridge the performance gap between large embodied foundation
models and compact, off-the-shelf VLMs, while substantially reducing the computational overhead associated with larger embodied foundation models.
We pursue this via a simple but principled hypothesis: compact VLMs already encode sufficient geometric and semantic information for spatial reasoning; a key bottleneck lies in the decoding strategy rather than insufficient representational capacity. Existing approaches either repurpose autoregressive text generation for coordinate regression, an inherently misaligned objective, or rely on heavyweight embodied foundation models that are impractical on resource-constrained hardware.
We instead keep the VLM \emph{entirely frozen} and attach a lightweight geometric prediction head that operates in its hidden feature space. The prediction head maps intermediate VLM representations to continuous spatial outputs using a DETR-style transformer decoder followed by task-specific feed-forward networks (Fig.~\ref{fig:overview}). We describe the model architecture and training strategy in the following subsections.

\subsection{Model Architecture}
\label{sec:arch}

Our model comprises two modules: (i) a \emph{frozen} VLM that serves as a
shared visual-linguistic encoder, and (ii) a lightweight \emph{geometry prediction head} consisting of a DETR-style transformer decoder and task-specific feed-forward networks (FFNs) that regress continuous spatial outputs. All learned parameters reside exclusively in the prediction head, keeping the
added footprint minimal.


\subsubsection{Vision-Language Model (VLM) Backbone}
\label{subsec:vlm}

We first instantiate an off-the-shelf VLM as a frozen backbone. We assume a canonical architecture: a visual encoder extracts patch features that are projected into the LLM embedding space via a learned adapter, yielding a sequence of visual token features $\mathbf{F}_I \in \mathbb{R}^{N_v \times d}$, where $N_v$ is the number of visual tokens and $d$ is the LLM hidden dimension. The text tokenizer maps the natural-language task instruction to token embeddings 
$\mathbf{F}_T \in \mathbb{R}^{N_t \times d}$. The two sequences are concatenated and processed by the LLM, which attends jointly over visual and textual context to produce a contextualized representation. 
Crucially, \emph{no weights of the VLM backbone are modified at any point}; improvements therefore arise from the learned geometric prediction interface operating on frozen VLM representations rather than from backbone adaptation.

\subsubsection{Geometry Prediction Head}
\label{subsec:head}

To decode geometric information latent in the frozen VLM, we adopt a DETR-style architecture \cite{carion2020end}. The frozen VLM backbone acts as the transformer encoder, and task-specific transformer decoder-FFN pairs independently predict bounding box coordinates (for affordance localization) and trajectory waypoints. This design avoids any autoregressive generation of coordinate tokens and instead treats geometric prediction as a direct regression problem, better aligned with the continuous and spatially structured nature of manipulation outputs.

\vspace{-2mm}
\paragraph{Transformer Decoder:}
Each decoder consists of $M$ layers, each comprising multi-head self-attention, multi-head cross-attention over the frozen visual memory tokens $\mathbf{V}$,
and a feed-forward sub-layer, with Pre-LN normalization throughout. We use a single learned query per head; the key design choice is how this query is initialized.

In autoregressive LLMs, the hidden state at the \emph{final text token position} at layer $l^*$, denoted $\mathbf{h}_{\mathrm{last}} \in \mathbb{R}^{d}$, attends to the full multimodal context and therefore encodes a holistic, instruction-grounded summary of the scene. We exploit this by conditioning the initial query on this representation:
\begin{equation}
    \mathbf{q}_0 = \mathbf{e}_{\mathrm{pos}} + W_c\, \mathbf{h}_{\mathrm{last}},
    \label{eq:query_init}
\end{equation}
where $\mathbf{e}_{\mathrm{pos}} \in \mathbb{R}^{d_{dec}}$ is a learned positional embedding and $W_c \in \mathbb{R}^{d_{dec} \times d}$ is a learned linear projection. This ensures that from the first cross-attention layer, the query is already biased toward image regions relevant to the specific instruction, rather than relying on the decoder to discover that bias from scratch.
The decoder then refines $\mathbf{q}_0$ across $M$ layers via cross-attention to $\mathbf{V}$, yielding a final representation $\mathbf{q}_M \in \mathbb{R}^{d_{\mathrm{dec}}}$ that is both semantically grounded in the instruction and spatially precise enough to decode into geometric coordinates.

\vspace{-2mm}
\paragraph{Prediction Feed-Forward Networks (FFNs):}
After $M$ decoder layers, the refined query $\mathbf{q}_M \in \mathbb{R}^{d_{dec}}$ is mapped to continuous spatial outputs by a task-specific prediction FFN: a 3-layer MLP with GeLU activations and hidden dimension $d_{dec}$. We maintain \emph{two independent decoder-FFN pairs}, one for affordance prediction and one for trajectory prediction. Both receive the same frozen VLM features as encoder context and the same conditioning signal $\mathbf{h}_{\mathrm{last}}$, but maintain separate weights and are trained with separate losses. This allows each head to specialize its cross-attention patterns and query refinement toward its respective geometric output, without interference between the two tasks.

\smallskip
\noindent\textbf{Affordance head:} The affordance FFN regresses to a normalized bounding box in corner format:
\begin{equation}
    \hat{b} =
        \mathrm{FFN}_{\mathrm{aff}}\!\left(\mathbf{q}^{\mathrm{aff}}_M\right)
        = \bigl(\hat{x}_1,\,\hat{y}_1,\,\hat{x}_2,\,\hat{y}_2\bigr).
    \label{eq:aff_output}
\end{equation}

\noindent\textbf{Trajectory head:} The trajectory FFN unrolls $T$ waypoints in a single forward
pass:
\begin{equation}
    \hat{\tau} = \mathrm{FFN}_{\mathrm{traj}}(\mathbf{q}^{\mathrm{traj}}_M)
    = \{(\hat{x}_t,\, \hat{y}_t)\}_{t=1}^{T}.
\end{equation}

In both cases, a sigmoid activation in the output layer constrains all predicted coordinates to $[0, 1]$, normalized with respect to the dimensions of the input image. At inference time, predictions are de-normalized back to pixel coordinates by multiplying by the original image width and height, respectively.

\subsection{Training}
\label{subsec:training}
\vspace{-2mm}
All training signals flow exclusively through the geometry prediction head; the VLM backbone remains frozen throughout. We define separate losses for the two tasks, each designed to handle the precise, spatial prediction for robot manipulation.

\subsubsection{Affordance Loss}



Given a predicted bounding box
$\hat{\mathbf{b}} = (\hat{x}_1, \hat{y}_1, \hat{x}_2, \hat{y}_2) \in [0,1]^4$ and ground-truth
$\mathbf{b} = (x_1, y_1, x_2, y_2) \in [0,1]^4$, we combine coordinate regression with a geometry-aware overlap term:
\begin{equation}
    \mathcal{L}_{\mathrm{afford}} =
        \lambda_{\mathrm{L1}}\,\mathcal{L}_{\mathrm{L1}}(\hat{\mathbf{b}}, \mathbf{b})
        + \lambda_{\mathrm{GIoU}}\,\mathcal{L}_{\mathrm{GIoU}}(\hat{\mathbf{b}}, \mathbf{b}).
    \label{eq:afford_loss}
\end{equation}

The coordinate term $\mathcal{L}_{\mathrm{L1}}$ is a Smooth L1 loss applied element-wise to the box corners, and the overlap term $\mathcal{L}_{\mathrm{GIoU}}$ is the Generalized IoU loss~\cite{rezatofighi2019generalized}. We use a tighter Smooth L1 threshold and standard GIoU; full definitions are provided in the supplementary material (Section~\ref{sec:supp_aff_loss}).

\subsubsection{Trajectory Loss}
\vspace{-2mm}
Given predicted waypoints $\hat{\mathcal{T}} = \{\hat{\mathbf{p}}_t\}_{t=1}^{T}$ and ground truth
$\mathcal{T} = \{\mathbf{p}_t\}_{t=1}^{T}$, with $\hat{\mathbf{p}}_t, \mathbf{p}_t \in [0,1]^2$, we define:
\begin{equation}
    \mathcal{L}_{\mathrm{traj}} =
        \lambda_{\mathrm{L1}}\,\mathcal{L}_{\mathrm{L1}}
        + \lambda_{\mathrm{start}}\,\mathcal{L}_{\mathrm{start}}
        + \lambda_{\mathrm{end}}\,\mathcal{L}_{\mathrm{end}}.
    \label{eq:traj_loss}
\end{equation}
The main term $\mathcal{L}_{\mathrm{L1}}$ is a Smooth L1 loss applied uniformly over all $T$ waypoints, and $\mathcal{L}_{\mathrm{start}}$ and $\mathcal{L}_{\mathrm{end}}$ are additional Smooth L1 penalties on the first and last waypoints that upweight errors at the contact and termination points of the trajectory. We use the same Smooth L1 defined in the affordance loss; full formulas are given in the supplementary material (Section ~\ref{sec:sup_traj_loss}).
\section{Experimental Analysis}
\label{sec: experiments}
\vspace{-2mm}
In this section we evaluate our method on both trajectory and affordance prediction and analyze the efficiency-accuracy trade-off against larger VLM baselines. 
We first discuss the experimental setup, including the baselines, datasets, evaluation metrics and implementation details. 

\subsection{Experimental Setup}
\subsubsection{Baselines} We evaluate our approach on two families of VLMs, Qwen2-VL \cite{wang2024qwen2}, using the 2B variety as our small VLM, and LLaVA-OneVision \cite{li2024llava-ov}, using the 0.5B variety. We compare against four categories of baseline methods:
\begin{itemize}
    \item \textbf{Baseline small VLMs:} We test both the compact Qwen2-VL-2B and LLaVA-OV-0.5B in their original decoding configurations: we generate coordinates autoregressively as text tokens with no task-specific training on ShareRobot \cite{ji2025robobrain}, representing the zero-shot geometric capability of off-the-shelf VLMs.
    \item \textbf{Baseline large VLMs:} We additionally test the larger 7B counterparts of our small VLMs (Qwen2-VL-7B and LLaVA-OV-7B) in their original decoding configurations, as a reference point for performance gains that can be unlocked with larger model capacity, despite incurring additional computational cost.
   \item \textbf{LoRA fine-tuned VLMs:} We fine-tune Qwen2-VL-2B and LLaVA-OV-0.5B on ShareRobot~\cite{ji2025robobrain} via next-token prediction on tokenized coordinates using LoRA~\cite{Hu2021LoRA:Models}. 
    \item \textbf{RoboBrain}~\cite{ji2025robobrain}: A VLM with large-scale training on embodied data. It is a generalist model that offers state-of-the-art performance across a range of robotic manipulation tasks, albeit at substantially greater computational cost.
\end{itemize}

\subsubsection{Datasets} We train on the ShareRobot dataset~\cite{ji2025robobrain}, which aggregates data from 23 sources in the Open-X Embodiment collection~\cite{o2024open}. For \emph{affordance prediction}, the training set contains 6{,}022 images and 500 validation images with bounding-box annotations $\{l_x, l_y, r_x, r_y\}$ denoting the affordance region for a given high-level task description. For \emph{trajectory prediction}, it contains 6{,}370 images and 500 validation images with gripper trajectories of up to $T=4$ waypoints $\{(x_t, y_t)\}_{t=1}^{T}$ in image space.

We evaluate the performance of our method and the baselines on two affordance benchmarks, AGD20K \cite{luo2022learning} and ShareRobot-Bench-Affordance \cite{ji2025robobrain}. We also evaluate on ShareRobot-Bench-Trajectory \cite{ji2025robobrain} for the trajectory prediction task. AGD20K provides out-of-distribution evaluation on diverse everyday objects, while ShareRobot-Bench assesses performance on manipulation-specific scenarios drawn from the same domain as the training data. Additional dataset and annotation details are provided in the Supplementary Section~\ref{sec:sup_agd_conversion}.

\subsubsection{Evaluation Metrics} Following prior work, for affordance prediction we report mean Average Precision (mAP$\uparrow$) across multiple IoU thresholds (0.25, 0.50, 0.75, 0.90) ~\cite{ji2025robobrain,tan2025roboos}.
For trajectory prediction, we report three complementary metrics \cite{ji2025robobrain, tan2025roboos}: Discrete Fréchet Distance (DFD$\downarrow$)~\cite{gu2023rt}, which measures order-sensitive trajectory shape; Hausdorff Distance (HD$\downarrow$)~\cite{sousa2020vehicle}, which captures the worst-case spatial gap; and Root Mean Squared Error (RMSE$\downarrow$), which measures average point-wise deviation.
Inference efficiency is reported as frames per second (FPS) and per-sample floating-point operations (TFLOPs). The detailed efficiency measurement protocol is described in Supplementary Section ~\ref{sec:efficiency_protocol}.

\subsubsection{Implementation Details} Our DETR-style transformer decoder comprises $M=3$ layers with hidden dimension $d_{dec}=256$ and 8 attention heads. The prediction FFN is a 3-layer MLP with hidden dimension 256 and GeLU activations. We optimize with AdamW~\cite{loshchilov2017decoupled} at a learning rate of $1\times10^{-4}$ with linear warmup followed by cosine decay, training for 20 epochs at batch size 16. Unless indicated otherwise, experiments are run on a single NVIDIA H100 GPU with bf16 mixed precision. Only the decoder and FFN heads are updated during training; the VLM backbone is frozen throughout. Implementation and training cost of LoRA fine-tuning can be found in Supplementary Section~\ref{sec:sup_implementation}.

\begin{table*}[t]
\centering
\caption{\textbf{Accuracy and inference efficiency comparison.}
    Our Geometric Head variants on frozen Qwen2-VL-2B and LLaVA-OV-0.5B backbones
    (\textit{Emb.}~=~\checkmark\ indicates embodied training; \textit{Head} $\in \{\text{token}, \text{geo}\}$).
    We report active backbone parameters (\#P(B)), accuracy metrics (DFD, HD, RMSE, mAP),
    throughput on H100 and RTX 3090 (FPS$\uparrow$), and compute cost (TFLOPs$\downarrow$).
    $^\dagger$Fully trained on embodied data; all other large VLMs are evaluated zero-shot.
    SRB=ShareRobot-Bench.
    Geo-Head substantially outperforms zero-shot and LoRA baselines and approaches
    7B/RoboBrain accuracy at a fraction of the parameters and compute.
}
\label{tab:combined_table}
\setlength{\tabcolsep}{1.5pt}
\scriptsize
\renewcommand{\arraystretch}{1.05}

\resizebox{\textwidth}{!}{%
\begin{tabular}{l c c c | c | c c c | c c c | c | c c | c c c}
\toprule
& & & & \multicolumn{7}{c|}{\textbf{Trajectory}} & \multicolumn{6}{c}{\textbf{Affordance}} \\
\cmidrule(lr){5-11} \cmidrule(lr){12-17}

\multirow{2}{*}{\textbf{Category}} &
\multirow{2}{*}{\textbf{Model}} &
\multirow{2}{*}{\textbf{Emb.}} &
\multirow{2}{*}{\textbf{Head}} &
\multirow{2}{*}{\textbf{\#P(B)}} &
\multicolumn{3}{c|}{\textbf{SRB}} &
\multicolumn{2}{c}{\textbf{FPS$\uparrow$}} &
\multirow{2}{*}{\textbf{TFLOPs$\downarrow$}} &
\multirow{2}{*}{\textbf{\#P(B)}} &
\textbf{AGD20K} &
\textbf{SRB} &
\multicolumn{2}{c}{\textbf{FPS$\uparrow$}} &
\multirow{2}{*}{\textbf{TFLOPs$\downarrow$}} \\

\cmidrule(lr){6-8}
\cmidrule(lr){9-10}
\cmidrule(lr){13-13}
\cmidrule(lr){14-14}
\cmidrule(lr){15-16}

& & & & &
\textbf{DFD$\downarrow$} &
\textbf{HD$\downarrow$} &
\textbf{RMSE$\downarrow$} &
\textbf{H100} &
\textbf{RTX} &
&
&
\textbf{mAP$\uparrow$} &
\textbf{mAP$\uparrow$} &
\textbf{H100} &
\textbf{RTX} &
\\

\midrule

\multirow{3}{*}{\shortstack{Large\\VLMs}}
  & RoboBrain$^\dagger$ \cite{ji2025robobrain}
  & \checkmark & token
  & 8.2
  & 0.191 & 0.171 & 0.133
  & 5.5 & 1.2 & 31.0
  & 8.2
  & 21.78 & 20.57
  & 4.1 & 0.8 & 80.0 \\

  & Qwen2-VL-7B \cite{wang2024qwen2}
  & \texttimes & token
  & 8.3
  & 0.602 & 0.591 & 0.494
  & 15.2 & 6.8 & 4.4
  & 8.3
  & 18.79 & 18.65
  & 6.2 & 1.7 & 22.0 \\

  & LLaVA-OV-7B \cite{li2024llava-ov}
  & \texttimes & token
  & 8.0
  & 0.641 & 0.591 & 0.494
  & 8.7 & 1.6 & 30.3
  & 8.0
  & 11.87 & 7.45
  & 4.1 & 0.8 & 80.0 \\

\midrule

  & Qwen2-VL-2B \cite{wang2024qwen2}
  & \texttimes & token
  & 2.2
  & 0.798 & 0.777 & 0.597
  & 16.6 & 11.1 & 1.63
  & 2.2
  & 7.30 & 6.27
  & 6.9 & 2.6 & 8.9 \\

  & \quad w/ LoRA
  & \checkmark & token
  & 2.22
  & $0.752 \pm 0.061$
  & $0.739 \pm 0.050$
  & $0.595 \pm 0.015$
  & 16.4 & 9.4 & 1.65
  & 2.22
  & $2.25 \pm 0.60$
  & $7.20 \pm 0.21$
  & 8.0 & 2.6 & 9.2 \\

\rowcolor{ourgreen}
  \cellcolor{white}
  & \quad w/ Geo-Head (Ours)
  & \checkmark & geo
  & 1.6
  & $0.251 \pm 0.010$
  & $0.219 \pm 0.008$
  & $0.177 \pm 0.007$
  & 21.0 & 13.8 & 1.2
  & 1.5
  & $12.37 \pm 1.16$
  & $11.58 \pm 0.58$
  & 7.9 & 3.0 & 6.7 \\

\cmidrule{2-17}

  & LLaVA-OV-0.5B \cite{li2024llava-ov}
  & \texttimes & token
  & 0.9
  & 0.777 & 0.768 & 0.650
  & 14.0 & 6.6 & 4.2
  & 0.9
  & 1.09 & 0.94
  & 7.7 & 3.2 & 11.8 \\

  & \quad w/ LoRA
  & \checkmark & token
  & 0.91
  & $0.695 \pm 0.001$
  & $0.700 \pm 0.003$
  & $0.581 \pm 0.002$
  & 13.6 & 6.4 & 4.4
  & 0.91
  & $2.39 \pm 0.11$
  & $2.51 \pm 1.10$
  & 7.7 & 3.2 & 12.0 \\

\rowcolor{ourgreen}
  \cellcolor{white}\multirow{-6}{*}{\shortstack{Small\\VLMs}}
  & \quad w/ Geo-Head (Ours)
  & \checkmark & geo
  & 0.8
  & $0.225 \pm 0.039$
  & $0.225 \pm 0.003$
  & $0.179 \pm 0.003$
  & 15.3 & 7.2 & 4.2
  & 0.7
  & $9.53 \pm 1.44$
  & $6.82 \pm 0.24$
  & 7.9 & 3.4 & 10.8 \\

\bottomrule
\end{tabular}%
}
\vspace{-2mm}
\end{table*}

\subsection{Trajectory and Affordance Prediction Results}
\label{sec:main_results}
\vspace{-2mm}
Table~\ref{tab:combined_table} presents the trajectory and affordance prediction performance across all models and baselines on ShareRobot-Bench and AGD20K, along with their inference speeds. 
Across both model families, our approach reduces trajectory DFD by over 68\% and improves affordance mAP by up to 8.7$\times$ over the corresponding vanilla small VLM baselines.
This supports our first claim that pairing small VLMs with geometrically compatible prediction heads enables substantial performance gains. These performance gains are accompanied by higher measured throughput on the H100 (7.9-21 FPS), as compared to 4.1-15.2 FPS for the larger VLM baselines. We discuss the performance of our method for the two tasks in more detail below.

\subsubsection{Trajectory Prediction} Our proposed geometry head yields the largest and most consistent gains on this task. Across both backbones, replacing autoregressive coordinate generation with direct regression achieves a relative reduction in DFD exceeding 60\%, with proportional reductions in HD and RMSE, confirming improvements in both global shape fidelity and point-wise accuracy. 
In particular, our Qwen2-VL-2B and LLaVA-OV-0.5B variants achieve DFDs of 0.251 and 0.225, respectively, approaching RoboBrain at 0.191 DFD, while achieving up to 3.8$\times$ faster inference on the H100 with 5--10$\times$ fewer parameters.

Together, these results validate our first claim that pairing small VLMs with a geometrically compatible prediction head enables substantial performance gains while maintaining superior efficiency over larger VLM counterparts.

LoRA fine-tuning of the base VLMs fails to deliver consistent improvements and remains substantially below our method across all metrics. For both backbones, fine-tuning only marginally reduces DFD, from 0.798 to $0.752 \pm 0.06$ for Qwen and from 0.777 to $0.695 \pm 0.001$ for LLaVA, while remaining substantially worse than the geometry-head variants. These results suggest that 
autoregressive text generation is a poorly aligned objective for continuous trajectory regression. 


\subsubsection{Affordance Prediction} Our geometry head yields strong gains on affordance prediction as well. Equipping the small VLMs with our geometry head increases affordance mAP for Qwen2-VL-2B by 1.7$\times$ (7.30 to $12.37 \pm 1.16$) and yields larger gains for LLaVA-OneVision-0.5B (1.09 to $9.53 \pm 1.44$), indicating that intermediate representations already encode rich spatial cues that token-based decoding fails to exploit. These gains come with equal or better runtime than the corresponding token heads and bring the small-geometry models close to, and in some cases competitive with, much larger 7B baselines, reinforcing our first claim that geometrically compatible prediction heads unlock strong affordance performance from compact VLMs while preserving their efficiency.




LoRA fine-tuning again fails to close the gap and in some cases actively degrades affordance quality relative to the zero-shot token baselines. Despite using the same backbones, LoRA-tuned models do not consistently improve over their frozen counterparts and remain substantially below our geometry-head variants, underscoring that simply pushing the language decoder harder
is a poor fit for precise spatial localization.

\subsubsection{Inference Efficiency}
\label{sec:efficiency}
Complementing the accuracy gains above, our geometry-head design preserves and often improves the runtime advantages of compact VLMs over their larger counterparts.

Table~\ref{tab:combined_table} reports inference efficiency for trajectory and affordance prediction on both an H100 and an RTX 3090, with the latter representing a lower-compute reference platform. A key structural advantage of our approach is that inference is truncated at the selected intermediate layer rather than executing the full VLM, reducing the active computation and FLOPs without modifying the backbone architecture. We provide a more detailed analysis of the efficiency--performance trade-off across layer indices in Supplementary Sec.~\ref{sec:sup_inf_eff_layer}. Importantly, the selected layer $l^*$ is backbone- and task-dependent; therefore, Qwen and LLaVA skip different fractions of the language-model computation. This leads to different FPS gains across models despite using the same geometric decoding strategy.

Across both tasks, geometry-head models generally achieve the highest throughput, with the 0.5B and 2B backbones attaining the fastest frame rates while operating at a fraction of the parameters and compute of their 7B counterparts. RoboBrain, while achieving the strongest overall performance, runs at substantially lower frame rates on an RTX 3090 (1.2/0.8 FPS), illustrating the substantially greater inference cost of the larger model. Together with the accuracy gains reported above, these results confirm that pairing small VLMs with a geometry-head delivers substantial performance improvements while maintaining superior efficiency over larger VLM counterparts.

\subsection{Disentangling Prediction Interface, Representation Depth and Decoder Architecture}
\label{sec:disentangle_abl}

\begin{table*}[t]
\centering
\caption{\textbf{Disentangling the source of performance gains.}
Continuous regression from the final VLM representation already substantially outperforms autoregressive token decoding, while intermediate features and DETR provide additional gains. Pre-LLM controls test performance before language-model contextualization. \textit{AR} denotes autoregressive decoding and \textit{LT+MLP} denotes direct regression from a learned token using an MLP. Results are reported as mean $\pm$ standard deviation where applicable.}
\label{tab:factorized_ablation}
\vspace{-2mm}
\setlength{\tabcolsep}{3pt}
\scriptsize
\renewcommand{\arraystretch}{1.08}

\resizebox{\textwidth}{!}{%
\begin{tabular}{l c c c c | cc | cc}
\toprule

\multirow{2}{*}{\textbf{Variant}} &
\multirow{2}{*}{\textbf{Representation}} &
\multirow{2}{*}{\shortstack{\textbf{Backbone}\\\textbf{Adaptation}}} &
\multirow{2}{*}{\textbf{Decoder}} &
\multirow{2}{*}{\textbf{Output}} &
\multicolumn{2}{c|}{\textbf{Qwen2-VL-2B}} &
\multicolumn{2}{c}{\textbf{LLaVA-OV-0.5B}} \\

\cmidrule(lr){6-7}
\cmidrule(lr){8-9}

& & & & &
\textbf{Traj. DFD$\downarrow$} &
\textbf{Aff. mAP$\uparrow$} &
\textbf{Traj. DFD$\downarrow$} &
\textbf{Aff. mAP$\uparrow$} \\

\midrule

Base VLM
& Full LLM
& None
& AR
& Tokens
& 0.798
& 7.30
& 0.777
& 1.09 \\

LoRA-AR
& Full LLM
& LoRA
& AR
& Tokens
& $0.752 \pm 0.061$
& $2.25 \pm 0.60$
& $0.695 \pm 0.001$
& $2.39 \pm 0.11$ \\

$l^*$-LoRA-AR
& Full LLM
& LoRA after $l^*$
& AR
& Tokens
& $0.680 \pm 0.010$
& $2.39 \pm 0.20$
& $0.680 \pm 0.030$
& $2.04 \pm 0.21$ \\

LoRA-Reg
& Final token
& LoRA
& LT+MLP
& Continuous
& $0.34 \pm 0.008$
& $3.24 \pm 0.06$
& $0.43 \pm 0.2$
& $5.78 \pm 1.04$ \\

\midrule

$l_0$-MLP
& $l_0$
& Frozen
& MLP
& Continuous
& $0.291 \pm 0.002$
& $8.74 \pm 0.97$
& $0.311 \pm 0.002$
& $8.68 \pm 0.10$ \\

$l_0$-DETR
& $l_0$
& Frozen
& DETR
& Continuous
& $0.265 \pm 0.002$
& $11.96 \pm 1.12$
& $0.293 \pm 0.001$
& $9.03 \pm 1.60$ \\

Final-MLP
& Final
& Frozen
& MLP
& Continuous
& $0.297 \pm 0.002$
& $7.92 \pm 0.54$
& $0.298 \pm 0.002$
& $9.06 \pm 0.14$ \\

Final-DETR
& Final
& Frozen
& DETR
& Continuous
& $0.289 \pm 0.037$
& $7.97 \pm 0.32$
& $0.225 \pm 0.039$
& $7.98 \pm 0.19$ \\

$l^*$-MLP
& $l^*$
& Frozen
& MLP
& Continuous
& $0.267 \pm 0.001$
& $9.36 \pm 0.03$
& $0.298 \pm 0.002$
& $9.16 \pm 1.30$ \\

\rowcolor{ourgreen}
$l^*$-DETR (Ours)
& $l^*$
& Frozen
& DETR
& Continuous
& $\mathbf{0.251 \pm 0.010}$
& $\mathbf{12.37 \pm 1.16}$
& $\mathbf{0.225 \pm 0.039}$
& $\mathbf{9.53 \pm 1.44}$ \\

\bottomrule
\end{tabular}%
}
\vspace{-2mm}
\end{table*}

To disentangle the source of the performance gains, we factorize the design into three components: the prediction interface (autoregressive tokens vs. direct continuous regression), representation depth, and decoder architecture (MLP vs. DETR) (see Table~\ref{tab:factorized_ablation}). 

For trajectory prediction, the prediction interface is the primary source of improvement. Decoding the final frozen VLM representation with a simple continuous MLP reduces DFD from 0.798 to $0.297\pm0.002$ for Qwen and from 0.777 to $0.298\pm0.002$ for LLaVA. This corresponds to 91.6\% and 86.8\% of the full improvement achieved by our $l^*$-DETR model, showing that most of the trajectory gain does not require either intermediate feature selection or a DETR decoder.

This interpretation is supported further by the LoRA controls. \emph{LoRA-AR} fine-tunes the VLM with LoRA while retaining autoregressive token generation, with $l^*$-\emph{LoRA-AR} applying LoRA only to layers after $l^*$. \emph{LoRA-Reg} replaces autoregressive generation with a learned regression token decoded by an MLP into continuous coordinates. As shown in Table~\ref{tab:factorized_ablation}, LoRA-Reg substantially improves over LoRA-AR for both VLM architectures. Conversely, retaining autoregressive prediction in $l^*$-\emph{LoRA-AR} remains substantially weaker than the continuous-regression variants. Together, these controls support the prediction interface as the dominant bottleneck for trajectory prediction.


Affordance prediction exhibits more backbone-specific behavior. For LLaVA, Final-MLP already recovers 94.4\% of the full mAP improvement, with $l^*$-MLP recovering 95.6\%, indicating that direct continuous prediction accounts for most of the gain. For Qwen, however, Final-MLP provides only a modest improvement, while moving the same MLP to $l^*$ recovers 40.6\% of the full gain; replacing the MLP with DETR at the same representation depth provides a further substantial improvement. Thus, for Qwen affordance prediction, representation depth and decoder architecture contribute more strongly alongside the prediction interface. The LoRA controls reinforce this interpretation: replacing autoregressive prediction with continuous regression improves affordance mAP when compared to LoRA-AR, while  $l^*$-LoRA-AR is comparable or worse, indicating that the prediction interface is important but does not alone explain the full affordance gains.

Overall, these results support our second claim: autoregressive decoding is the dominant bottleneck for trajectory prediction. For affordance prediction, it is a major bottleneck which can have stronger backbone-dependent interaction with representation depth and decoder architecture.

\begin{figure}[t]
    \centering
    \includegraphics[width=0.8\linewidth]{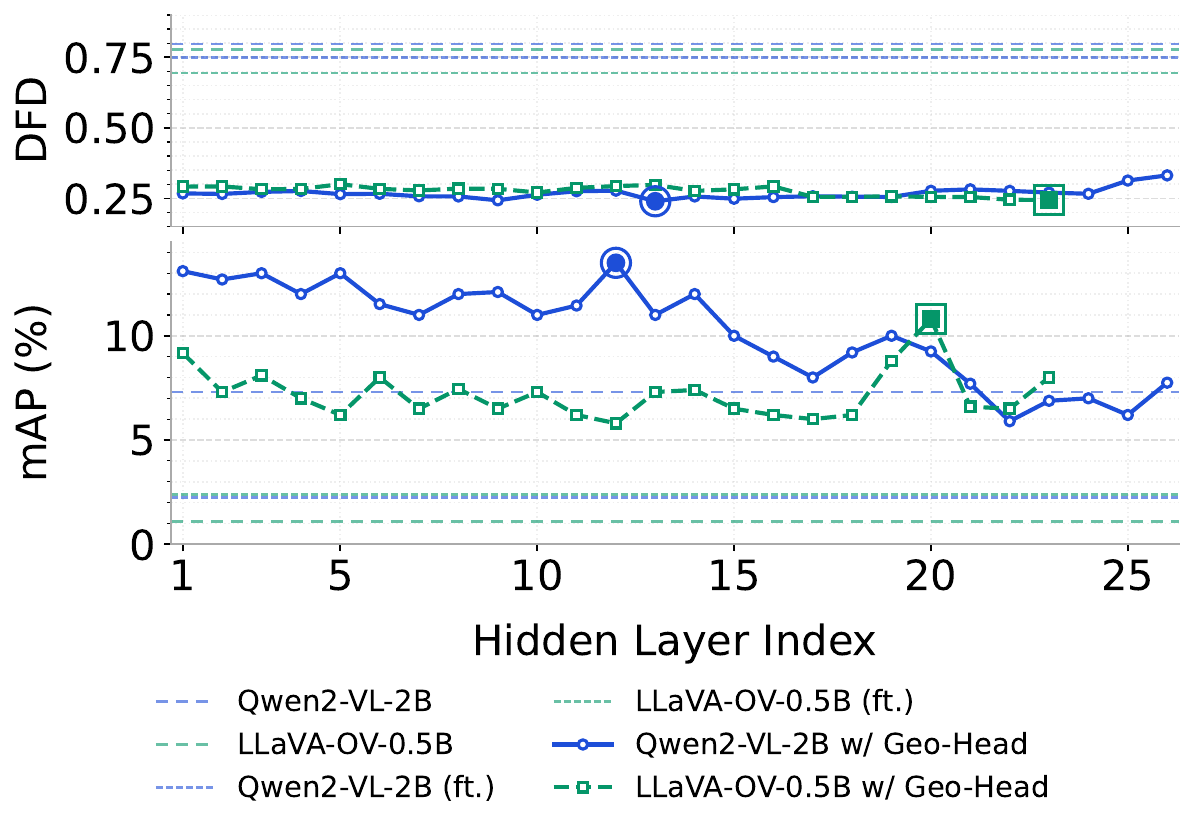}
    \caption{\textbf{Layer-wise representational analysis.} Trajectory DFD on ShareRobot-Bench-Trajectory and affordance mAP on AGD20K across all language-model layers for Qwen2-VL-2B and LLaVA-OV-0.5B, compared against the base VLM and LoRA fine-tuned autoregressive baselines. Colored markers indicate the selected layer used in the final model.}
    \label{fig:layer_sweep}
    \vspace{-2mm}
\end{figure}

\subsection{Accessibility of Spatial Information Across VLM Representations}
\label{sec:spatial_information}
Having established that direct continuous prediction can expose substantial spatial information from frozen VLM representations, we next examine how the accessibility of this information varies across representation depth. The pre-LLM (\(l_0\)) controls in Table 2 demonstrated that strong geometric prediction is already possible from the projected multimodal features before any LLM transformer layer, indicating that the LLM does not create the spatial signal from scratch. We then attach the same geometry head to hidden states at different VLM depths to examine how prediction quality changes across layers.

Figure~\ref{fig:layer_sweep} presents a full layer sweep across both backbones and tasks, together with the base VLM and LoRA fine-tuned autoregressive baselines. For each backbone, we select $l^*$ independently for each task: the layer with minimum trajectory error ($l=13$ for Qwen and $l=20$ for LLaVA) and highest affordance mAP ($l=12$ for Qwen and $l=23$ for LLaVA).

For trajectory prediction, every evaluated hidden layer yields substantially lower DFD than the corresponding autoregressive baselines for both backbones. Together with the strong pre-LLM result, this suggests that trajectory-relevant spatial information is already broadly accessible throughout the VLM representation hierarchy, but is not effectively exposed by next-token coordinate prediction. 

Affordance prediction is more depth-dependent. The geometry head outperforms the autoregressive baselines across most layers, but the magnitude of the improvement varies substantially with depth. Combined with the strong pre-LLM performance, this suggests that affordance-relevant spatial information is already present early in the representation hierarchy, while subsequent layers can make it more or less accessible for geometric decoding.

Overall, compact VLMs contain substantial spatial information from the input to the LLM onward. Its accessibility is relatively stable across depth for trajectory prediction, but more strongly representation-dependent for affordance prediction. Intermediate LLM layers can refine or reorganize this information, but deeper representations are not necessarily better suited to precise geometric prediction. 


We additionally ablate two design choices in the geometric head: decoder-query conditioning and explicit start/endpoint supervision. The preferred query representation is backbone- and task-dependent, while endpoint supervision consistently improves trajectory accuracy and reduces collapse toward a dataset-level mean. Full results are provided in Supplementary Sec.~\ref{sec:supp_add_abl}.

\subsection{Qualitative Results}

\begin{figure}[t]
    \centering
    \setlength{\tabcolsep}{1pt}
    \renewcommand{\arraystretch}{0}
    \begin{tabular}{cccc}
        \multicolumn{2}{c}{\footnotesize\textbf{Qwen2-VL 2B}} &
        \multicolumn{2}{c}{\footnotesize\textbf{LLaVA-OV 0.5B}} \\[1pt]

        \includegraphics[width=0.23\columnwidth, height=1.5cm]{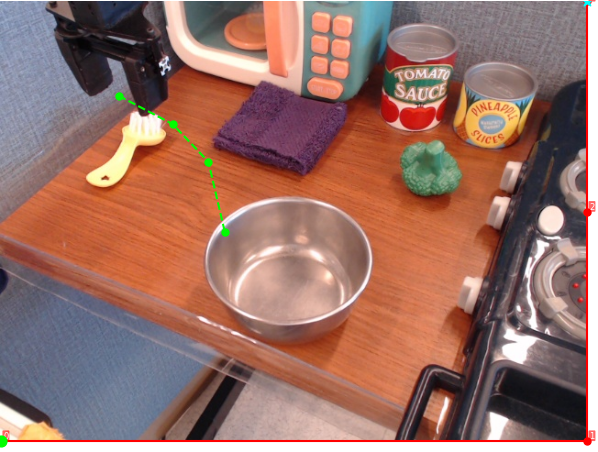} &
        \includegraphics[width=0.23\columnwidth, height=1.5cm]{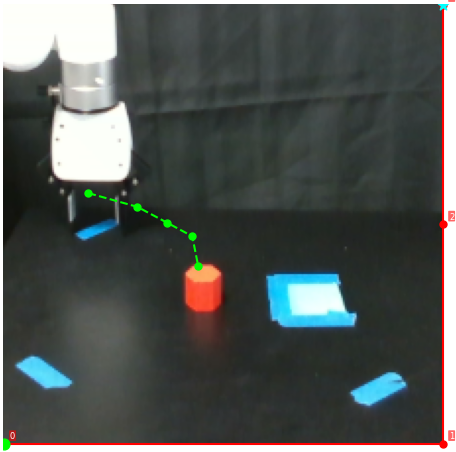} &
        \includegraphics[width=0.23\columnwidth, height=1.5cm]{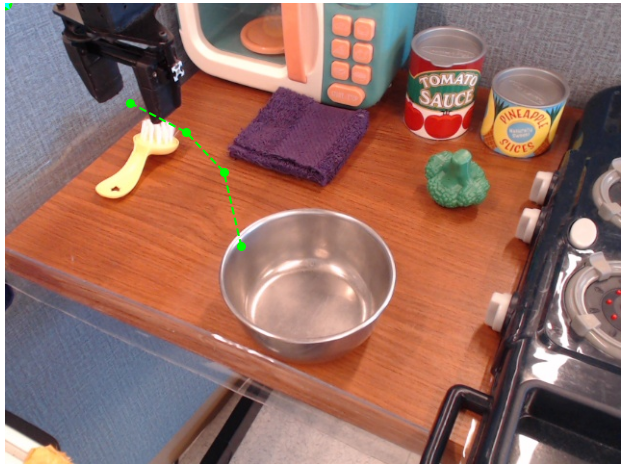} &
        \includegraphics[width=0.23\columnwidth, height=1.5cm]{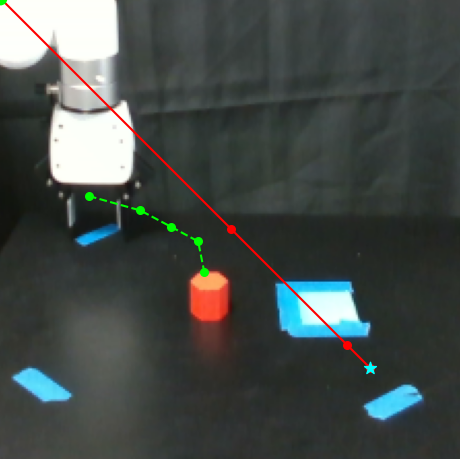} \\[2pt]

        \multicolumn{2}{c}{\footnotesize\textbf{+ Geo-Head (Ours)}} &
        \multicolumn{2}{c}{\footnotesize\textbf{+ Geo-Head (Ours)}} \\[1pt]

        \includegraphics[width=0.23\columnwidth, height=1.5cm]{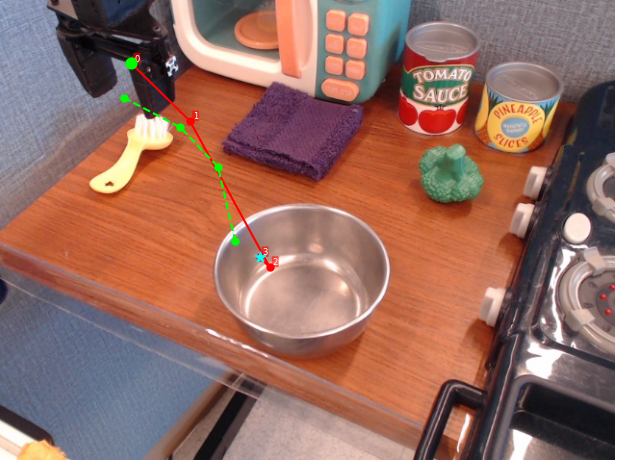} &
        \includegraphics[width=0.23\columnwidth, height=1.5cm]{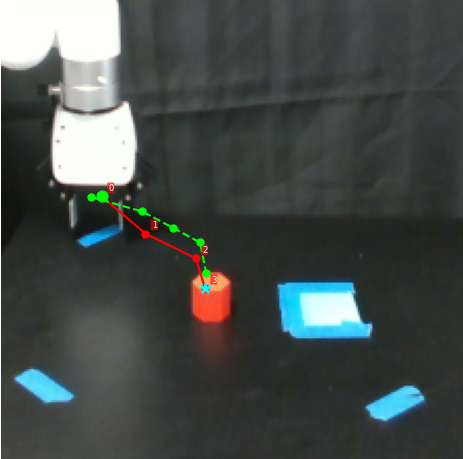} &
        \includegraphics[width=0.23\columnwidth, height=1.5cm]{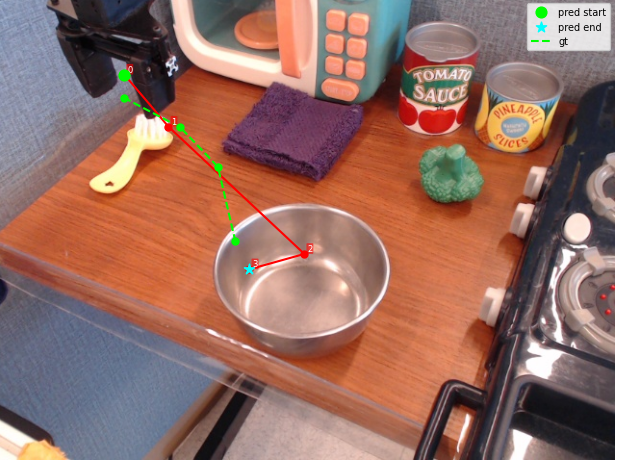} &
        \includegraphics[width=0.23\columnwidth, height=1.5cm]{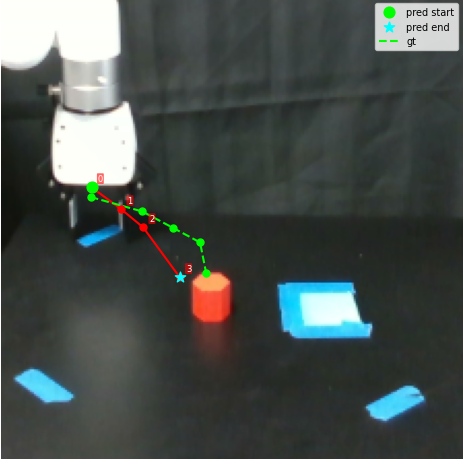} \\[1pt]
    \end{tabular}

    \caption{\textbf{Qualitative trajectory predictions.} Red: predicted, green: ground truth. Top row: vanilla VLM baseline. Bottom row: with our geometry head.}
    \label{fig:qualitative_traj}
    \vspace{-2mm}
\end{figure}

\begin{figure}[t]
    \centering
    \setlength{\tabcolsep}{1pt}
    \renewcommand{\arraystretch}{0}
    \begin{tabular}{cccc}
        \multicolumn{2}{c}{\footnotesize\textbf{Qwen2-VL 2B}} &
        \multicolumn{2}{c}{\footnotesize\textbf{LLaVA-OV 0.5B}} \\[1pt]

        \includegraphics[width=0.23\columnwidth, height=1.5cm]{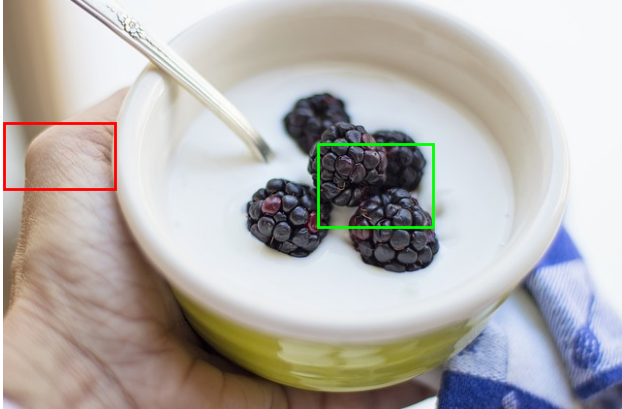} &
        \includegraphics[width=0.23\columnwidth, height=1.5cm]{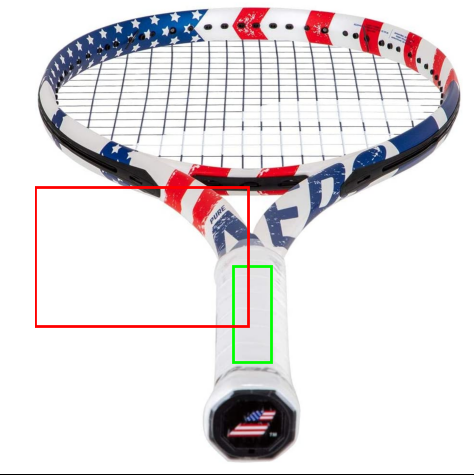} &
        \includegraphics[width=0.23\columnwidth, height=1.5cm]{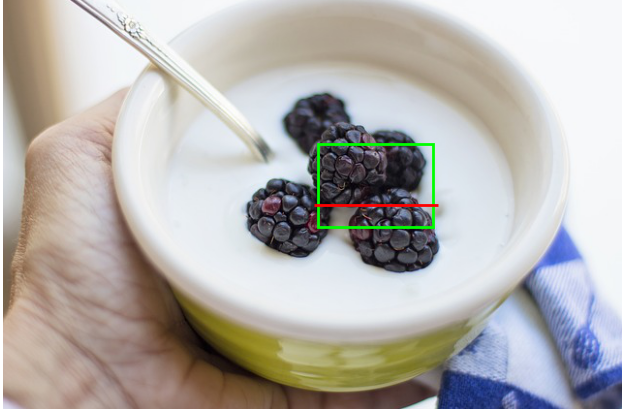} &
        \includegraphics[width=0.23\columnwidth, height=1.5cm]{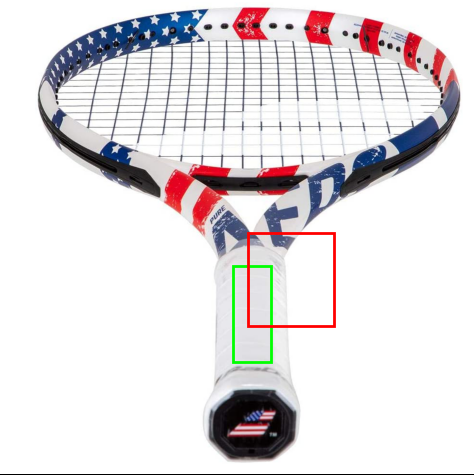} \\[2pt]

        \multicolumn{2}{c}{\footnotesize\textbf{+ Geo-Head (Ours)}} &
        \multicolumn{2}{c}{\footnotesize\textbf{+ Geo-Head (Ours)}} \\[1pt]

        \includegraphics[width=0.23\columnwidth, height=1.5cm]{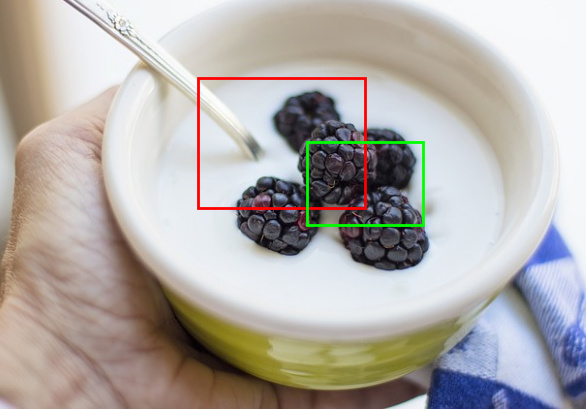} &
        \includegraphics[width=0.23\columnwidth, height=1.5cm]{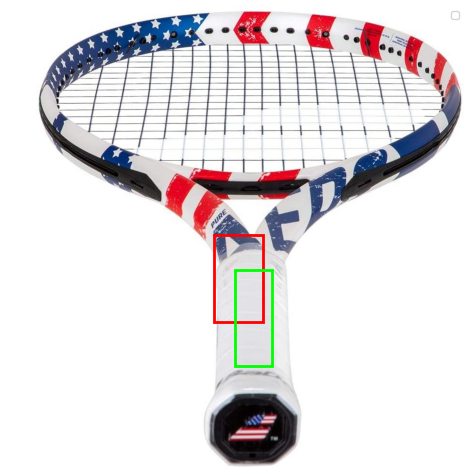} &
        \includegraphics[width=0.23\columnwidth, height=1.5cm]{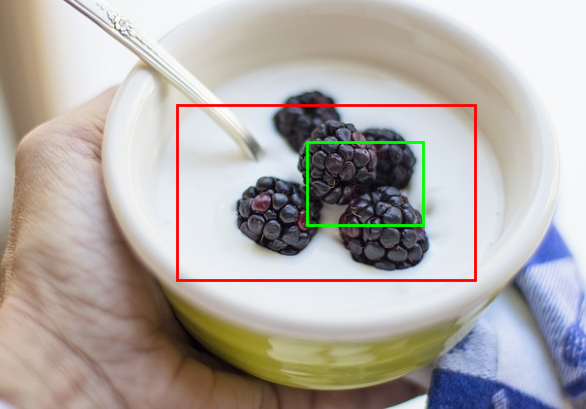} &
        \includegraphics[width=0.23\columnwidth, height=1.5cm]{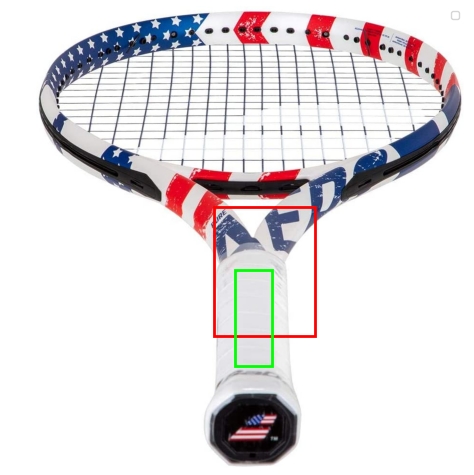} \\[1pt]
    \end{tabular}

    \caption{\textbf{Qualitative affordance predictions.} Red: predicted box, green: ground truth. Top row: vanilla VLM baseline. Bottom row: with our geometry head.}
    \label{fig:qualitative_aff}
    \vspace{-2mm}
\end{figure}

Figures \ref{fig:qualitative_traj} and \ref{fig:qualitative_aff} complement the quantitative results. For trajectory prediction, vanilla VLMs produce fragmented, spatially inconsistent waypoints, whereas the geometry head yields coherent trajectories with well-anchored endpoints. For affordance prediction, vanilla VLMs often roughly localize the relevant interaction region, suggesting that useful spatial information is already present but not precisely exposed. The geometry head turns these representations into tighter, more coherent predictions across both tasks, visually supporting our core claims. Additional qualitative results are provided in Supplementary Sec.~\ref{sec:supp_qual}.

\section{Conclusion}
\vspace{-3pt}

Compact VLMs of 0.5B--2B parameters already encode useful spatial information across their representation hierarchy, although its accessibility varies with task and model depth. We show that direct continuous prediction is the primary source of improvement for trajectory prediction, while affordance prediction depends more strongly on the interaction between prediction interface, representation depth, and decoder architecture. Building on this, we attach a lightweight geometry-compatible prediction head to a frozen, off-the-shelf VLM and directly regress affordance boxes and trajectory waypoints without modifying the backbone. This interface improves both tasks while preserving compact-VLM efficiency. Our 2B model approaches \mbox{RoboBrain} on trajectory prediction while running 3.8$\times$ faster (21 vs. 5.5 FPS) and requiring substantially less computation (1.2 vs. 31 TFLOPs). These results suggest a broader principle for efficient robot perception: before scaling the backbone, first ensure that the prediction interface is aligned with the geometric structure of the task.


Three limitations motivate future work. First, affordance and trajectory are predicted with separate heads; a unified decoder could better exploit their geometric interdependence. Second, early exit reduces active computation but still requires storing the backbone up to $l^*$; quantization or distillation into a truncated model could further reduce deployment cost. Third, our evaluation is limited to instruction-conditioned 2D geometric prediction prior to execution. Future work should extend the approach to 3D and closed-loop robotic settings and evaluate downstream manipulation success.

\section*{Acknowledgment}
The authors acknowledge support from CSIRO’s Data61 Embodied AI Cluster and partial support from the Australian Research Council (ARC) Discovery Project Grant DP250103634. D.M. acknowledges ongoing support from the QUT Centre for Robotics. We also acknowledge the Queensland University of Technology (QUT) for subsidised access to its High Performance Computing (HPC) facilities, and the QUT eResearch team for their expertise in enabling this project.
{
    \small
    \bibliographystyle{ieeenat_fullname}
    \bibliography{main}
}
\clearpage
\setcounter{page}{1}
\appendix
\twocolumn[{
\begin{center}
    {\large \textbf{Decoding Geometry, Not Tokens: Efficient VLMs for Robot Affordance and Trajectory Prediction} }\\[0.5em]
    {\large Supplementary Material}

\end{center}
\vspace{1em}
}]

\section{Training}
\subsection{Affordance Loss}
\label{sec:supp_aff_loss}
\paragraph{Smooth L1 coordinate loss.}
Given predicted and ground-truth boxes
$\hat{\mathbf{b}}, \mathbf{b} \in [0,1]^4$,
the coordinate regression term is

\begin{equation}
    \mathcal{L}_{\mathrm{L1}}(\hat{\mathbf{b}}, \mathbf{b}) =
        \frac{1}{4}\sum_{i=1}^{4}
        \ell_{\beta}\!\left(\hat{b}_i - b_i\right),
\end{equation}

where the Smooth L1 function $\ell_{\beta}$ is defined as

\begin{equation}
    \ell_{\beta}(x) =
    \begin{cases}
        \dfrac{x^2}{2\beta} & \text{if } |x| < \beta, \\[6pt]
        |x| - \dfrac{\beta}{2} & \text{otherwise.}
    \end{cases}
\end{equation}

We set $\beta = 0.05$, making the loss substantially more sensitive to small coordinate errors than the standard $\beta = 1$ formulation; this is important for precise affordance localization in manipulation tasks.

\paragraph{GIoU overlap loss.}
The overlap term is the Generalized IoU loss~\cite{rezatofighi2019generalized}:

\begin{equation}
    \mathcal{L}_{\mathrm{GIoU}}(\hat{\mathbf{b}}, \mathbf{b}) =
        1 - \left[
            \mathrm{IoU}(\hat{\mathbf{b}}, \mathbf{b})
            - \frac{|\mathcal{C} \setminus (\hat{\mathbf{b}} \cup \mathbf{b})|}{|\mathcal{C}|}
        \right],
\end{equation}

where $\mathcal{C}$ is the smallest axis-aligned enclosing box of $\hat{\mathbf{b}}$ and $\mathbf{b}$.

\subsection{Trajectory Loss}
\label{sec:sup_traj_loss}
Given predicted waypoints $\hat{\mathcal{T}} = \{\hat{\mathbf{p}}_t\}_{t=1}^{T}$ and ground truth
$\mathcal{T} = \{\mathbf{p}_t\}_{t=1}^{T}$, with $\hat{\mathbf{p}}_t, \mathbf{p}_t \in [0,1]^2$, the same Smooth L1 is uniformly applied over all $T$ waypoints, with $\beta=0.05$ making the loss more sensitive to small coordinate errors than the standard $\beta = 1$ formulation:
\begin{equation}
    \mathcal{L}_{\mathrm{L1}} =
        \frac{1}{T} \sum_{t=1}^{T}
        \ell_{\beta}\!\left(\hat{\mathbf{p}}_t - \mathbf{p}_t\right),
\end{equation}

Because the first and last waypoints determine where contact is made and where the manipulation terminates, we add explicit endpoint penalties:
\begin{equation}
    \mathcal{L}_{\mathrm{start}} = \ell_{\beta}(\hat{\mathbf{p}}_{1} - \mathbf{p}_{1}),
    \qquad
    \mathcal{L}_{\mathrm{end}}   = \ell_{\beta}(\hat{\mathbf{p}}_{T} - \mathbf{p}_{T}),
\end{equation}
which upweight the contribution of the start and end of the trajectory relative to intermediate waypoints. 

\section{AGD20K Affordance Box Construction}
\label{sec:sup_agd_conversion}
AGD20K does not provide ground-truth bounding boxes for affordance prediction. We therefore construct box annotations from the evaluation-set affordance masks, similar to \cite{ji2025robobrain}. For each binary mask, Connected Component Analysis (CCA) identifies contiguous non-zero regions, from which bounding boxes are generated. Box extent is controlled by a threshold during mask-to-box conversion, with larger values producing tighter boxes. We use a threshold of $0.7$, which produced stable annotations across the evaluation set. These boxes serve as ground-truth labels and are evaluated using the same bounding-box protocol as in the main paper.

\section{Inference Efficiency Protocol}
\label{sec:efficiency_protocol}


We measure inference efficiency on NVIDIA H100 and RTX 3090 GPUs with batch size 1. Qwen2-VL uses dynamic-resolution preprocessing, while LLaVA-OneVision uses aspect-ratio-preserving resizing and high-resolution image-grid processing; the same preprocessing is used across variants of each backbone. Model loading is excluded, and FPS is measured from preprocessed inputs to final predictions after 20 warm-up iterations, averaged over 1000 samples.

FLOPs are computed with PyTorch Profiler over 10 samples using the same input resolution and model configuration. For early-exit variants, FLOPs include computation through $l^*$ and the geometric head, excluding later layers. All models follow the same profiling procedure.

\section{Implementation Details}
\label{sec:sup_implementation}
The LoRA baselines use rank 16 with $\alpha=8$ and are applied to all linear projection layers in the LLM. They contain 18.46M and 8.80M trainable parameters for Qwen2-VL-2B and LLaVA-OV-0.5B, respectively, compared with 4.08M for the geometric head. Training to convergence on a single NVIDIA H100 GPU required approximately 15\,h for LoRA, compared with 4\,h for the geometric head.

\section{Efficiency--Performance Trade-off Across Exit Depth}
\label{sec:sup_inf_eff_layer}

\begin{figure}[t]
    \centering
    \includegraphics[width=0.75\linewidth]{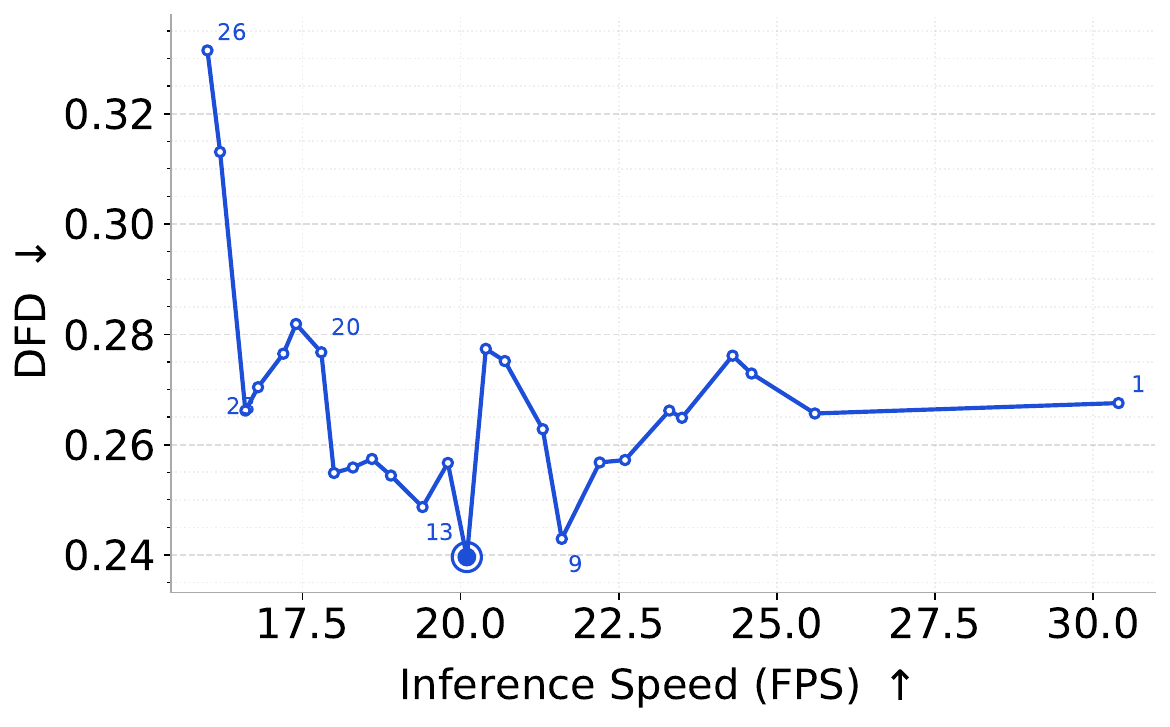}
    \vspace{-2mm}   
    \caption{Efficiency–performance trade-off across layers on Qwen2-VL-2B for the trajectory prediction task.}
    \label{fig:sup_layer_efficiency}
\end{figure}

Figure~\ref{fig:sup_layer_efficiency} shows the trajectory accuracy--efficiency trade-off for Qwen2-VL-2B across different exit layers. Earlier exits increase throughput, while deeper exits reduce FPS without consistently improving accuracy. The best DFD is obtained at $l^*=13$, which retains substantially higher throughput than the deepest exits. These results show that intermediate-layer decoding is not only beneficial for representation quality, but also provides a favorable accuracy--efficiency trade-off by avoiding unnecessary later-layer computation.

\section{Additional Ablation Studies}
\label{sec:supp_add_abl}
\subsection{Query Conditioning Strategy}
We compare two strategies for conditioning the DETR decoder query: the hidden state of the \emph{final text token} at layer $l^*$, and \emph{attention pooling} over task instruction tokens. As shown in Table~\ref{tab:query_token}, the impact differs across architectures: attention pooling improves trajectory prediction on Qwen2-VL-2B but degrades LLaVA-OV-0.5B, while the final text token is consistently stronger for affordance prediction across both backbones. To maintain a consistent decoder design across architectures and tasks, we use the final text token as the query-conditioning representation in all main-paper experiments.

\begin{table}[t]
\centering
\caption{Ablation of query token selection for trajectory prediction on ShareRobot-Bench-Trajectory and affordance prediction on AGD20K.}
\label{tab:query_token}
\resizebox{\columnwidth}{!}{%
\setlength{\tabcolsep}{4pt}

\begin{tabular}{l l ccc c}
\toprule
& & \multicolumn{3}{c}{\textbf{Trajectory}} & \textbf{Affordance} \\
\cmidrule(lr){3-5} \cmidrule(lr){6-6}
\textbf{Backbone} & \textbf{Query}
& \textbf{DFD$\downarrow$} & \textbf{HD$\downarrow$} & \textbf{RMSE$\downarrow$}
& \textbf{mAP$\uparrow$} \\
\midrule
\multirow{2}{*}{Qwen2-VL-2B}
& Last token & $0.251 \pm 0.010$ & $0.219\pm 0.008$ & $0.177 \pm 0.007$ & \textbf{$12.37 \pm 1.16$} \\
& Attn. pool & \textbf{$0.215 \pm 0.002$} & \textbf{$0.194 \pm 0.001$} & \textbf{$0.155 \pm 0.002$} & $8.67 \pm 1.01$ \\
\midrule
\multirow{2}{*}{LLaVA-OV-0.5B}
& Last token & \textbf{$0.225 \pm 0.039$} & \textbf{$0.225 \pm 0.003$} & \textbf{$0.179 \pm 0.003$} & \textbf{$9.53 \pm 1.44$} \\
& Attn. pool & $0.256 \pm 0.004$ & $0.24 \pm 0.01$ & $0.189 \pm 0.002$ & $7.27 \pm 1.2$ \\
\bottomrule
\end{tabular}%
}
\end{table}

\begin{table}[t]
\centering
\caption{Ablation on start and end point supervision for trajectory prediction head (Qwen2-VL-2B, ShareRobot-Bench).}
\label{tab:traj_ablation_startpoint}
\resizebox{\columnwidth}{!}{%
\setlength{\tabcolsep}{6pt}
\begin{tabular}{l ccc}
\toprule
\textbf{Model} & \textbf{DFD↓} & \textbf{HD↓} & \textbf{RMSE↓} \\
\midrule
w/ start \& end point loss & \textbf{$0.251 \pm 0.010$} & \textbf{$0.219 \pm 0.008$} & \textbf{$0.177 \pm 0.007$} \\
w/ only overall point loss     & $0.322 \pm 0.02$          & $0.26 \pm 0.01$          & $0.26 \pm 0.03$          \\
\bottomrule
\end{tabular}%
}
\end{table}

\subsection{Trajectory Loss Components}
Table~\ref{tab:traj_ablation_startpoint} shows that explicitly supervising the start and end points of predicted trajectories yields consistent improvements across all metrics (DFD: $0.251 \pm 0.010$ vs.\ $0.322 \pm 0.02$, HD: $0.219 \pm 0.008$  vs. $0.26 \pm 0.01$, RMSE: $0.177 \pm 0.007$ vs.\ $0.26 \pm 0.03$). Without these endpoint terms, the geometry head tends to collapse to a dataset-level mean trajectory, whereas adding start/end supervision restores spatially diverse, input-dependent predictions. This improvement shows that the decoding objective strongly influences precise geometric prediction even with a frozen backbone.

\label{sec:sup_abl}
\begin{figure}[ht]
    \centering
    \begin{subfigure}[b]{0.35\linewidth}
        \includegraphics[width=\linewidth]{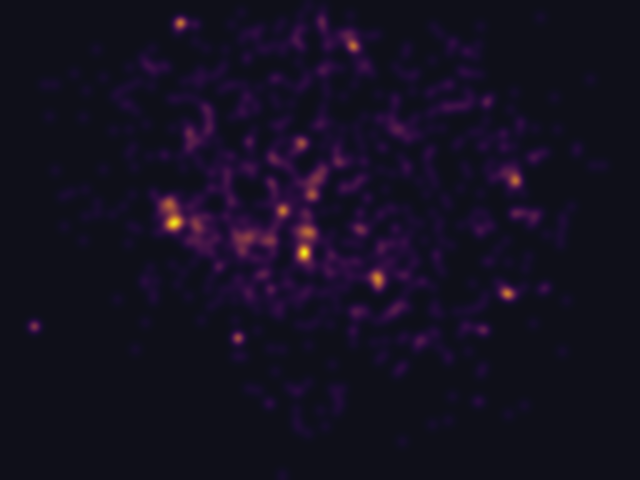}
        \caption{}
        \label{fig:heatmap_a}
    \end{subfigure}
    \begin{subfigure}[b]{0.35\linewidth}
        \includegraphics[width=\linewidth]{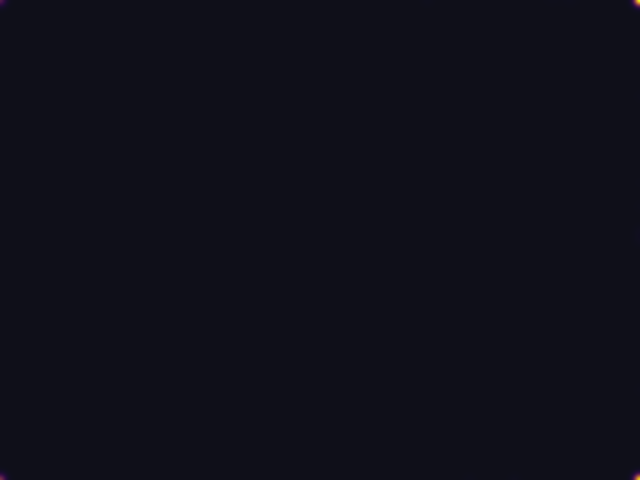}
        \caption{}
        \label{fig:heatmap_b}
    \end{subfigure}
    \begin{subfigure}[b]{0.35\linewidth}
        \includegraphics[width=\linewidth]{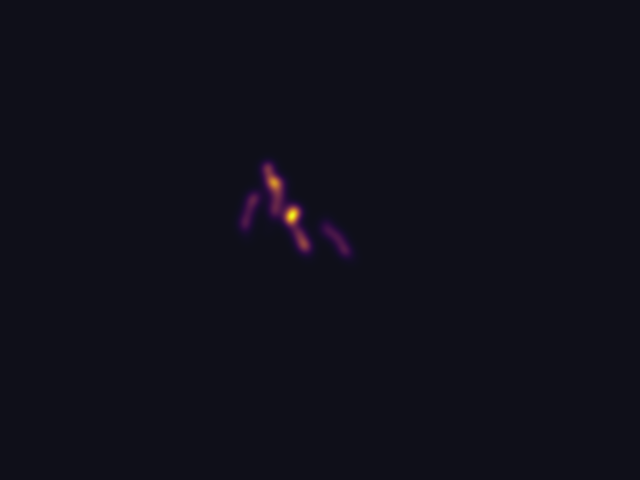}
        \caption{}
        \label{fig:heatmap_c}
    \end{subfigure}
    \begin{subfigure}[b]{0.35\linewidth}
        \includegraphics[width=\linewidth]{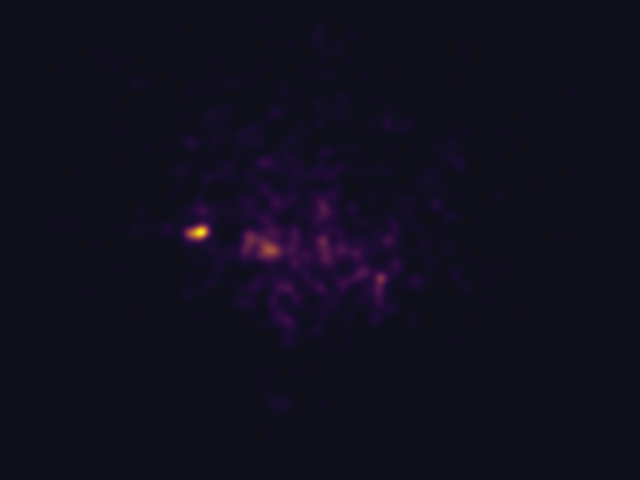}
        \caption{}
        \label{fig:heatmap_d}
    \end{subfigure}
    \caption{Predicted waypoint heatmaps over the test set, visualizing how endpoint supervision shapes the spatial distribution of trajectory predictions. \textbf{(a)} Ground-truth distribution: waypoints are broadly spread, reflecting the spatial diversity of manipulation trajectories in the dataset. \textbf{(b)} Zero-shot Qwen2-VL-2B baseline: predictions collapse to the image boundaries, confirming that vanilla autoregressive generation fails to produce spatially meaningful trajectories. \textbf{(c)} Geometry head \emph{without} endpoint supervision: predictions collapse to a single central region, as the uniform L1 objective is minimised by regressing toward the dataset mean. \textbf{(d)} Geometry head \emph{with} endpoint supervision: the distribution recovers spatial diversity matching the ground truth, demonstrating that boundary supervision forces input-specific localization rather than dataset-level averaging.}
    \vspace{-2mm}
    \label{fig:heatmap}
\end{figure}




The heatmaps in Figure~\ref{fig:heatmap} illustrate how endpoint supervision shapes spatial prediction behavior. Without it, the geometry head collapses to the image center — the uniform L1 objective is minimized by regressing toward the dataset mean. Adding explicit start and end point supervision breaks this collapse and recovers the spatially diverse distribution seen in the ground truth.

\section{Qualitative Results}
\begin{figure*}[ht]
    \centering
    \setlength{\tabcolsep}{1pt}
    \renewcommand{\arraystretch}{0}
    \begin{tabular}{ccc}
        \includegraphics[width=0.32\textwidth, height=4cm]{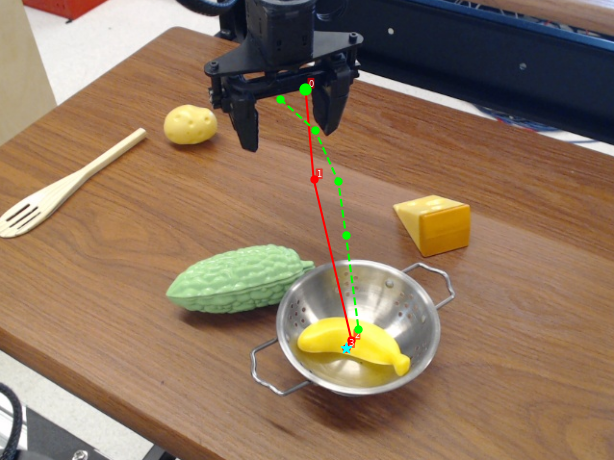} &
        \includegraphics[width=0.32\textwidth, height=4cm]{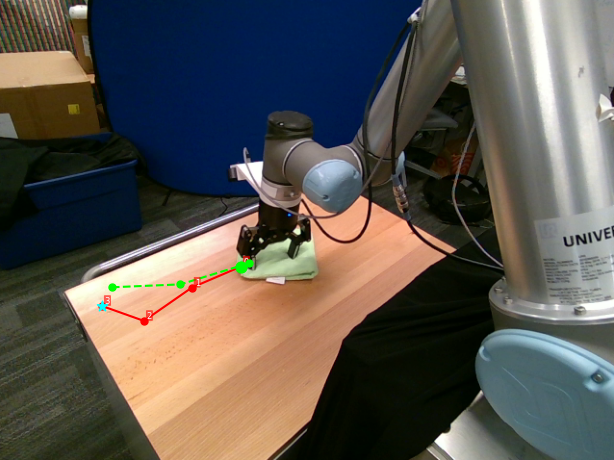} &
        \includegraphics[width=0.32\textwidth, height=4cm]{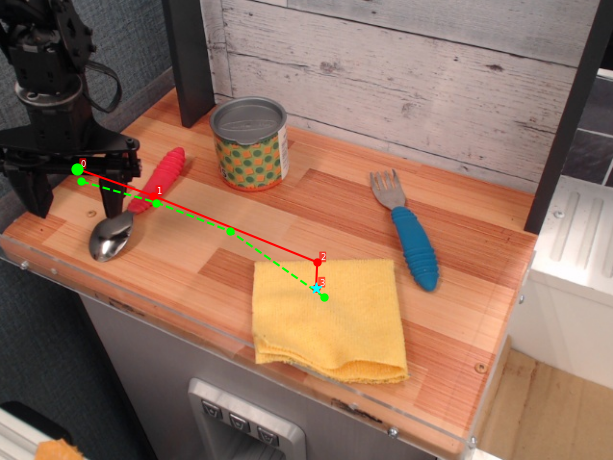} \\ [3pt]
        {\small Approach banana} & {\small Drag cloth left} & {\small Reach for cloth} \\[7pt]
        \includegraphics[width=0.32\textwidth, height=4cm]{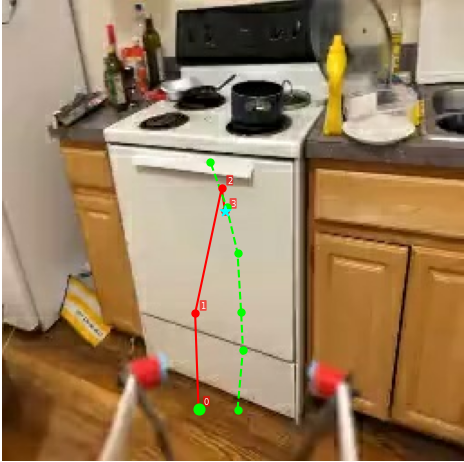} &
        \includegraphics[width=0.32\textwidth, height=4cm]{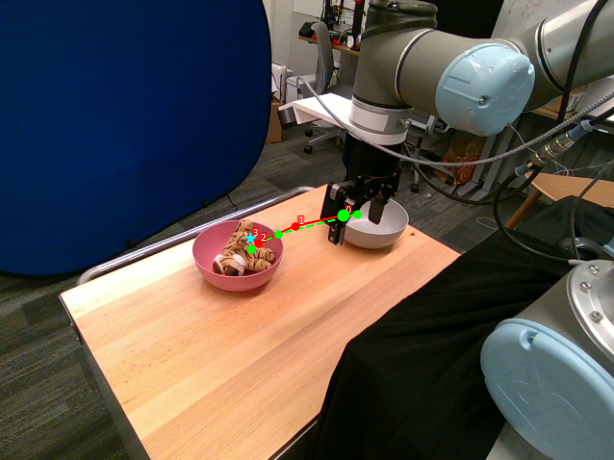} &
        \includegraphics[width=0.32\textwidth, height=4cm]{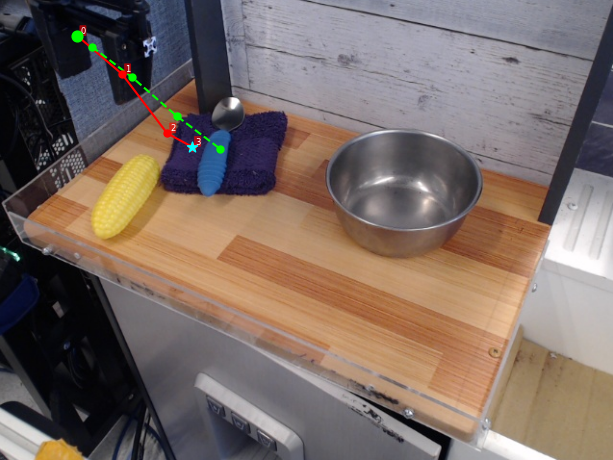} \\ [3pt]
        {\small Approach oven} & {\small Reach stuffed animal inside bowl} & {\small Reach for spoon} \\[7pt]
        \includegraphics[width=0.32\textwidth, height=4cm]{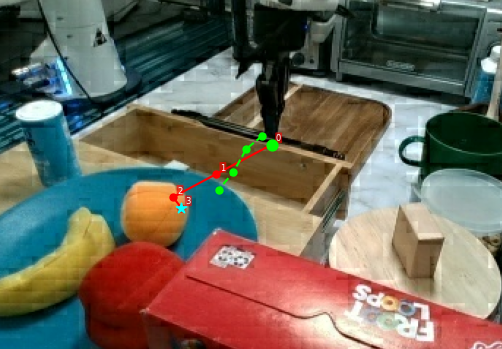} &
        \includegraphics[width=0.32\textwidth, height=4cm]{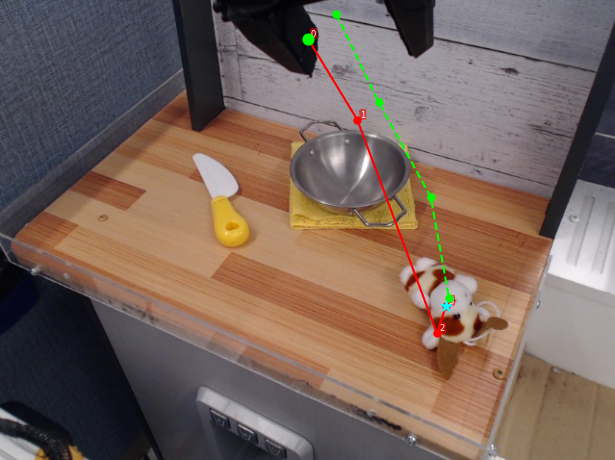} &
        \includegraphics[width=0.32\textwidth, height=4cm]{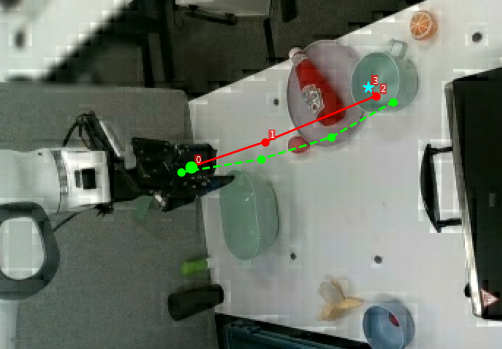} \\ [3pt]
        {\small Close drawer} & {\small Reach stuffed animal} & {\small Reach for mug} \\[7pt]
        \midrule \\[7pt]
        \includegraphics[width=0.32\textwidth, height=4cm]{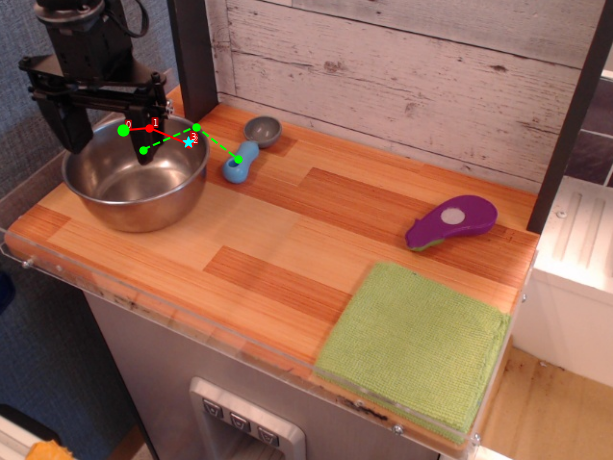} &
        \includegraphics[width=0.32\textwidth, height=4cm]{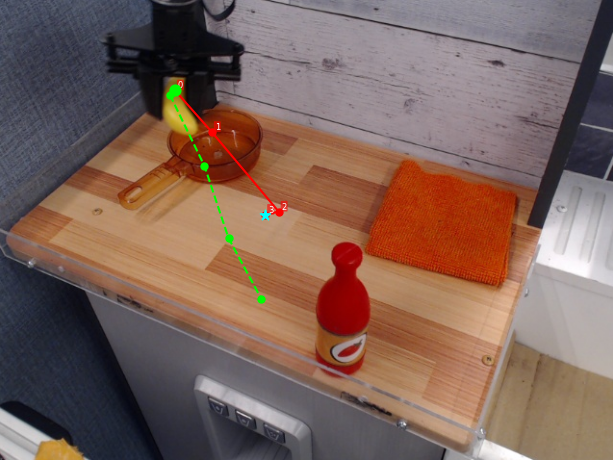} &
        \includegraphics[width=0.32\textwidth, height=4cm]{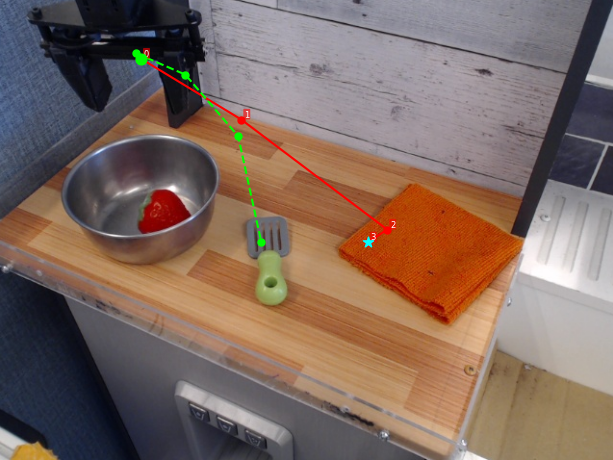} \\ [3pt]
        {\small Reach for blue spoon} & {\small Place banana next to bottle} & {\small Reach for spatula} \\
    \end{tabular}
    \caption{\textbf{Additional visualizations of diverse trajectories predicted by Qwen2-VL-2B + Geo-Head.} Green curve represents the ground-truth and the red curve represents the prediction. The visualizations in the first three rows demonstrate that our geometry head effectively generates end-effector manipulation trajectories based on the observation and task instruction. The fourth row presents some failure cases arising from object-localization errors or insufficient spatial understanding.}
    \label{fig:qwen_traj}
\end{figure*}

\begin{figure*}[t]
    \centering
    \setlength{\tabcolsep}{1pt}
    \renewcommand{\arraystretch}{0}
    \begin{tabular}{ccc}
        \includegraphics[width=0.32\textwidth, height=4cm]{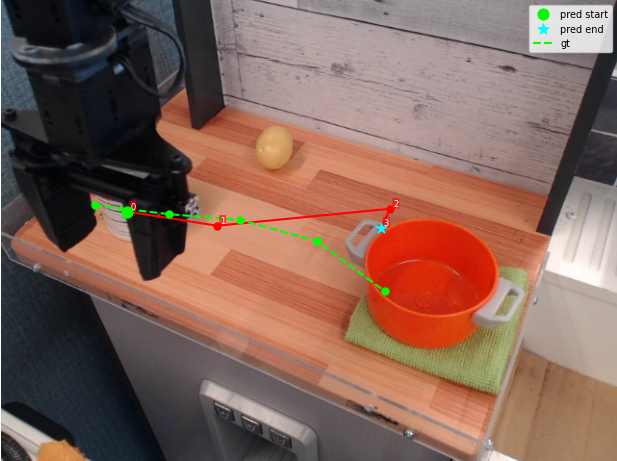} &
        \includegraphics[width=0.32\textwidth, height=4cm]{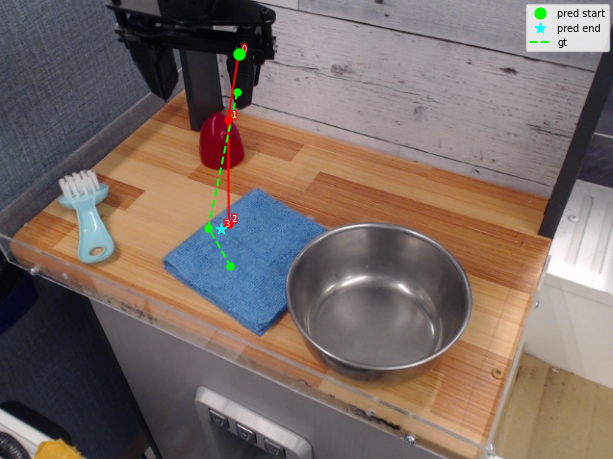} &
        \includegraphics[width=0.32\textwidth, height=4cm]{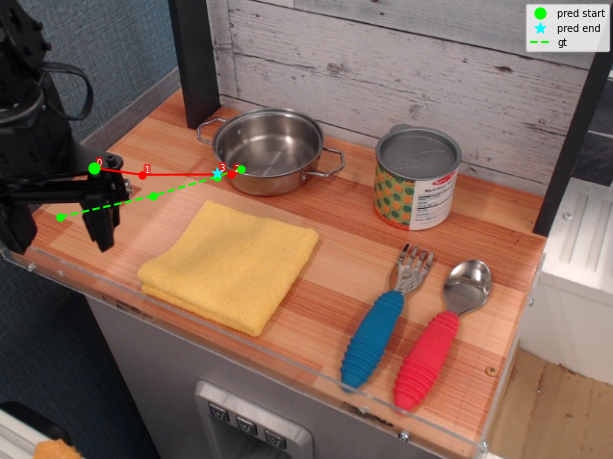} \\ [3pt]
        {\small Lower gripper towards pot} & {\small Reach for cloth} & {\small Reach for pot} \\[7pt]
        \includegraphics[width=0.32\textwidth, height=4cm]{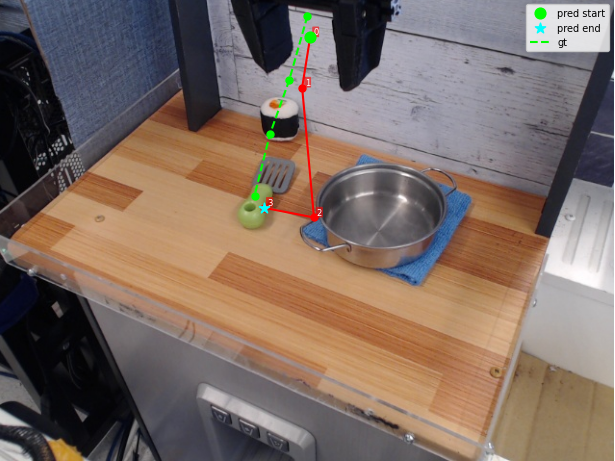} &
        \includegraphics[width=0.32\textwidth, height=4cm]{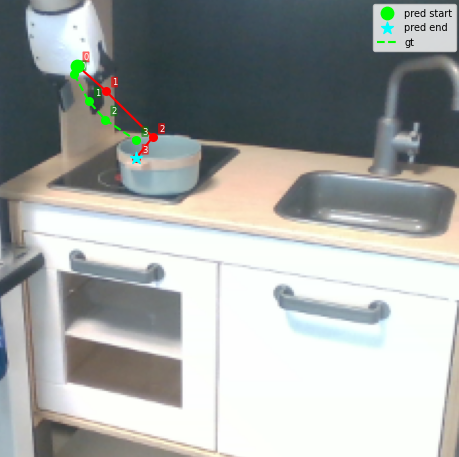} &
        \includegraphics[width=0.32\textwidth, height=4cm]{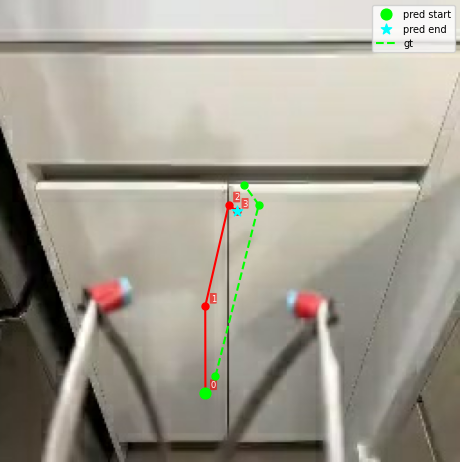} \\ [3pt]
        {\small Reach for spatula} & {\small Reach for pot} & {\small Reach for cabinet doors} \\[7pt]
        \includegraphics[width=0.32\textwidth, height=4cm]{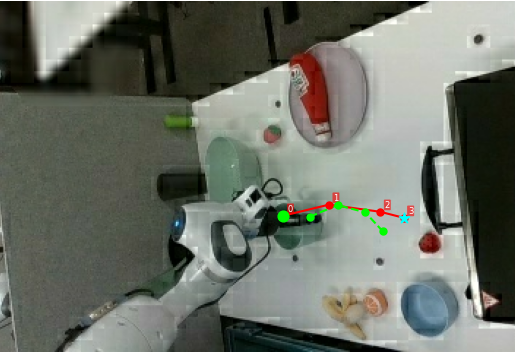} &
        \includegraphics[width=0.32\textwidth, height=4cm]{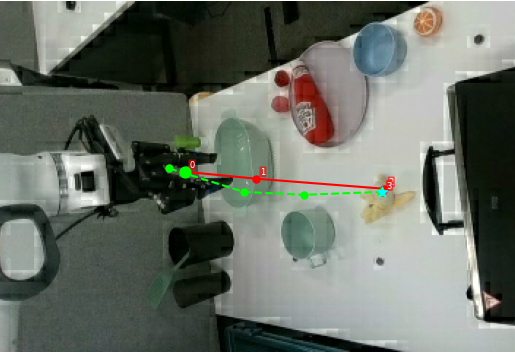} &
        \includegraphics[width=0.32\textwidth, height=4cm]{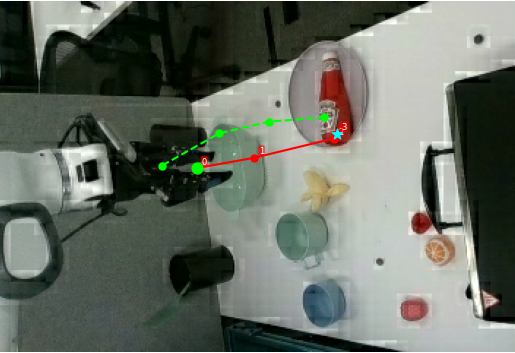} \\ [3pt]
        {\small Drag mug to the right} & {\small Reach for the banana} & {\small Reach for ketchup} \\[7pt]
        \midrule \\[7pt]
        \includegraphics[width=0.32\textwidth, height=4cm]{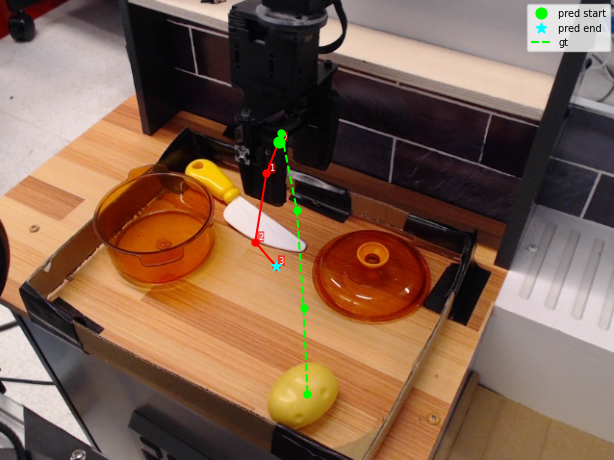} &
        \includegraphics[width=0.32\textwidth, height=4cm]{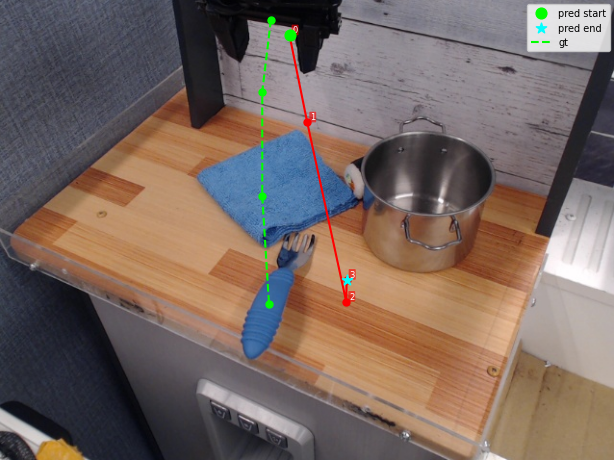} &
        \includegraphics[width=0.32\textwidth, height=4cm]{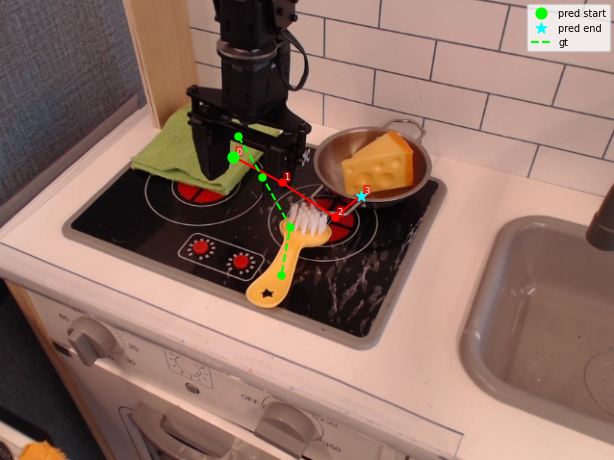} \\ [3pt]
        {\small Reach for potato} & {\small Reach for fork} & {\small Reach for yellow brush} \\
    \end{tabular}
    \caption{\textbf{Additional visualizations of diverse trajectories predicted by LLaVA-OV-0.5B + Geo-Head.} Green curve represents the ground-truth and the red curve represents the prediction. The visualizations in the first three rows demonstrate that our geometry head effectively generates end-effector manipulation trajectories based on the observation and task instruction. The fourth row presents some failure cases.}
    \label{fig:llava_traj}
\end{figure*}

\begin{figure*}[t]
    \centering
    \setlength{\tabcolsep}{1pt}
    \renewcommand{\arraystretch}{0}
    \begin{tabular}{ccc}
        \includegraphics[width=0.32\textwidth, height=4cm]{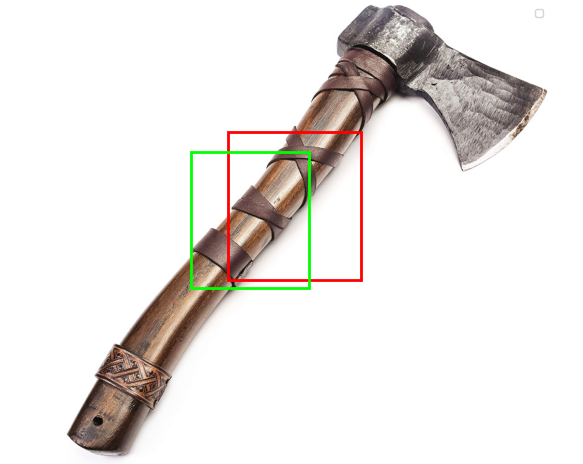} &
        \includegraphics[width=0.32\textwidth, height=4cm]{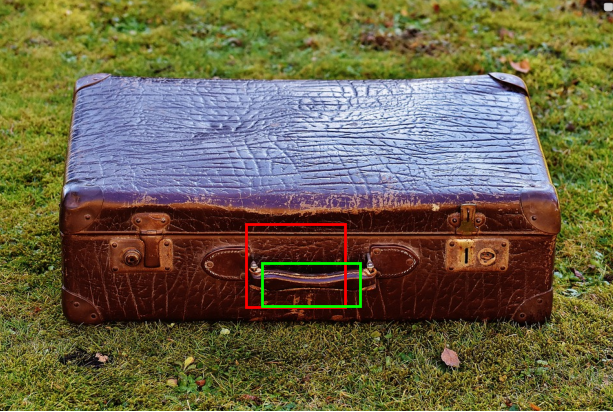} &
        \includegraphics[width=0.32\textwidth, height=4cm]{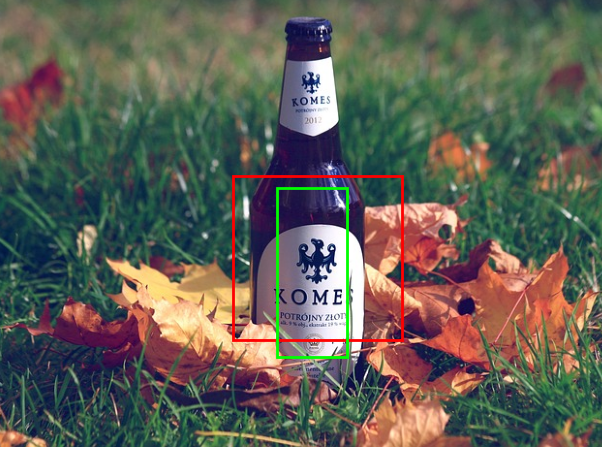} \\ [3pt]
        {\small Hold axe} & {\small Pick up suitcase} & {\small Hold bottle} \\[7pt]
        \includegraphics[width=0.32\textwidth, height=4cm]{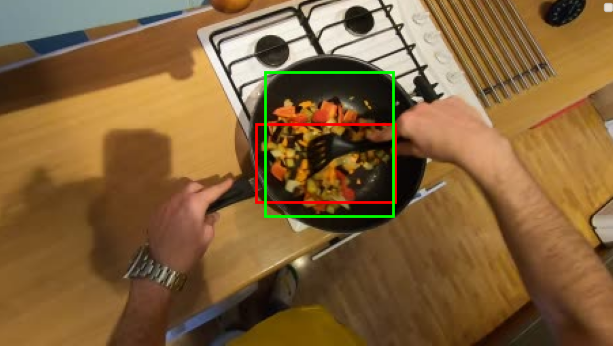} &
        \includegraphics[width=0.32\textwidth, height=4cm]{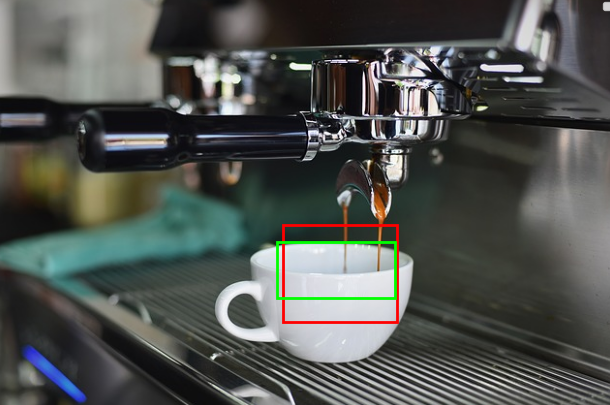} &
        \includegraphics[width=0.32\textwidth, height=4cm]{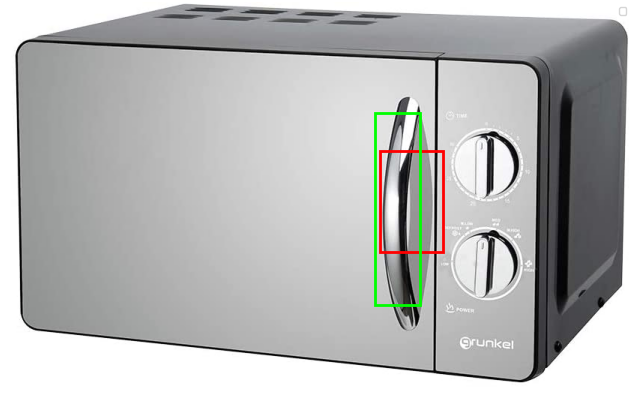} \\ [3pt]
        {\small Stir pan} & {\small Drink with cup} & {\small Open microwave} \\[7pt]
        \includegraphics[width=0.32\textwidth, height=4cm]{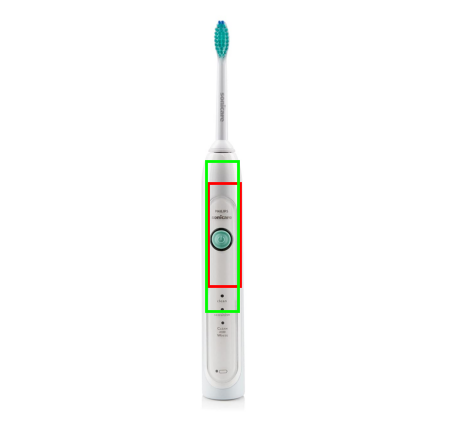} &
        \includegraphics[width=0.32\textwidth, height=4cm]{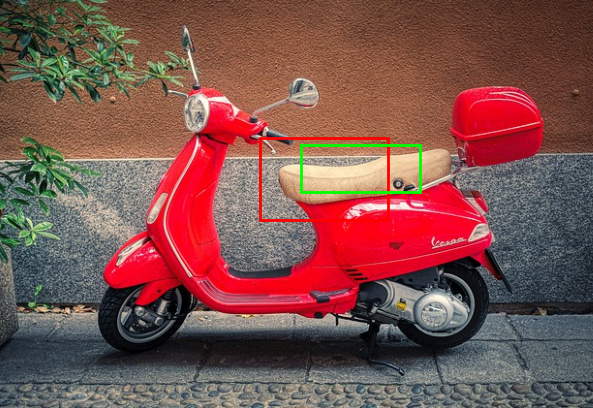} &
        \includegraphics[width=0.32\textwidth, height=4cm]{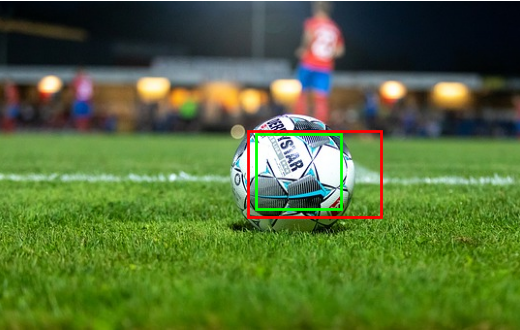} \\ [3pt]
        {\small Hold toothbrush} & {\small Sit on bike}& {\small Catch ball} \\[7pt]
        \midrule \\[7pt]
        \includegraphics[width=0.32\textwidth, height=4cm]{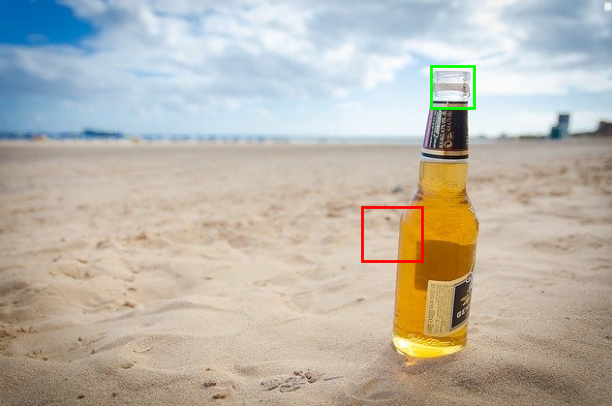} &
        \includegraphics[width=0.32\textwidth, height=4cm]{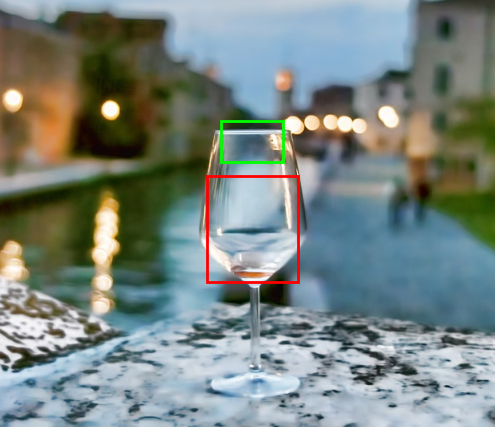} &
        \includegraphics[width=0.32\textwidth, height=4cm]{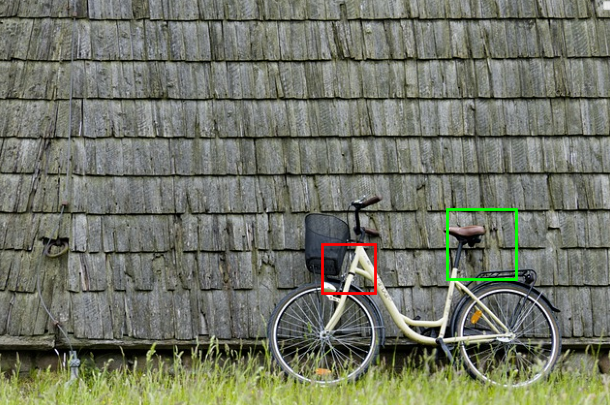} \\ [3pt]
        {\small Drink from bottle} & {\small Sip wine glass} & {\small Sit on bicycle} \\
    \end{tabular}
    \caption{\textbf{Additional visualizations of diverse affordance regions predicted by Qwen2-VL-2B + Geo-Head.} Green box represents the ground-truth and the red box represents the prediction. The visualizations in the first three rows demonstrate that our geometry head effectively generates affordance boxes based on the observation and task instruction. The fourth row presents some failure cases. }
    \label{fig:qwen_aff}
\end{figure*}

\begin{figure*}[t]
    \centering
    \setlength{\tabcolsep}{1pt}
    \renewcommand{\arraystretch}{0}
    \begin{tabular}{ccc}
        \includegraphics[width=0.32\textwidth, height=4cm]{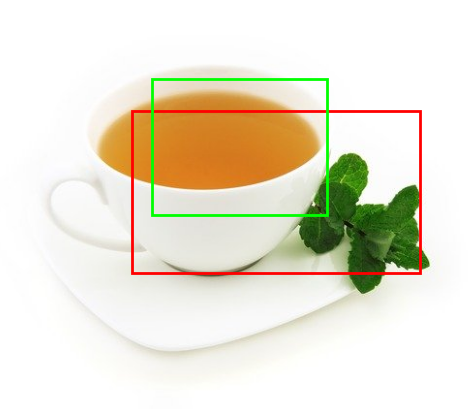} &
        \includegraphics[width=0.32\textwidth, height=4cm]{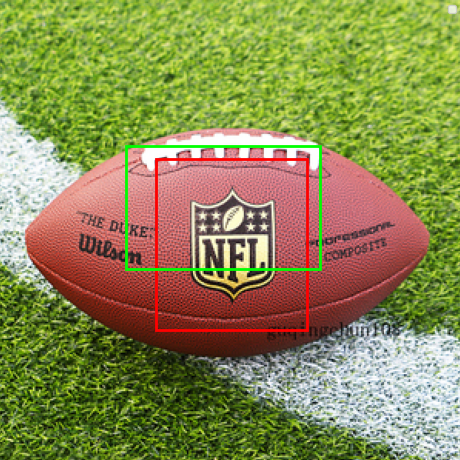} &
        \includegraphics[width=0.32\textwidth, height=4cm]{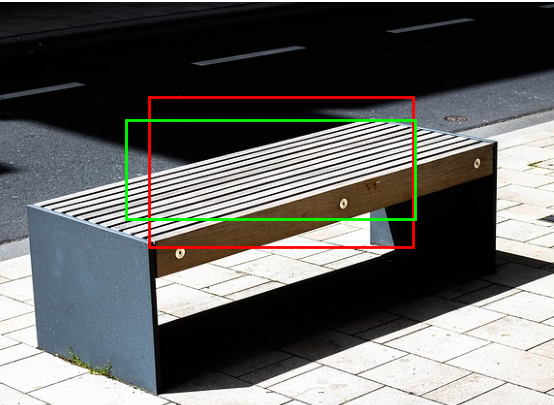} \\ [3pt]
        {\small Drink from cup} & {\small Catch rugby ball} & {\small Lie on bench} \\[7pt]
        \includegraphics[width=0.32\textwidth, height=4cm]{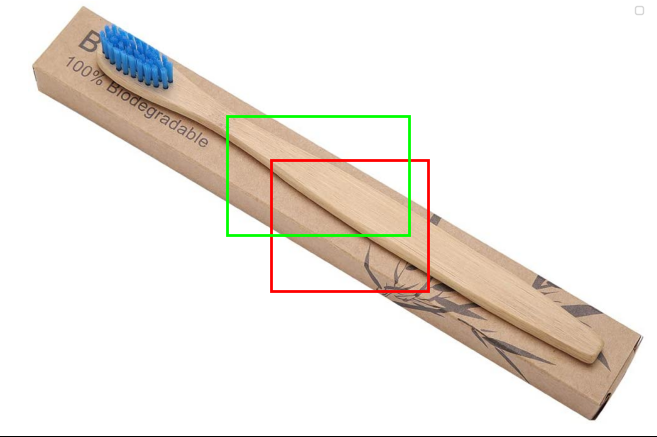} &
        \includegraphics[width=0.32\textwidth, height=4cm]{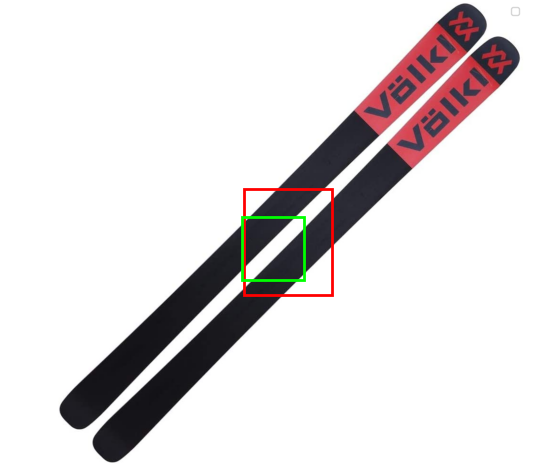} &
        \includegraphics[width=0.32\textwidth, height=4cm]{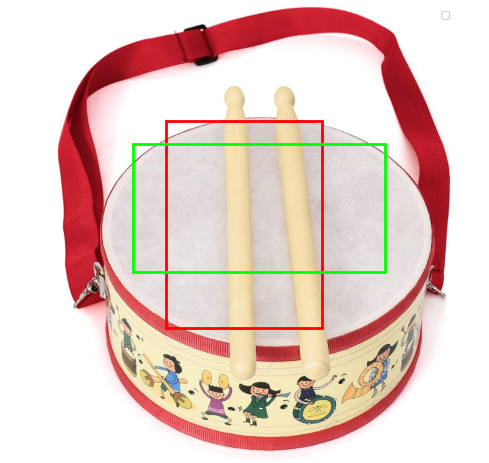} \\ [3pt]
        {\small Hold toothbrush} & {\small Pick up skis} & {\small Beat drum} \\[7pt]
        \includegraphics[width=0.32\textwidth, height=4cm]{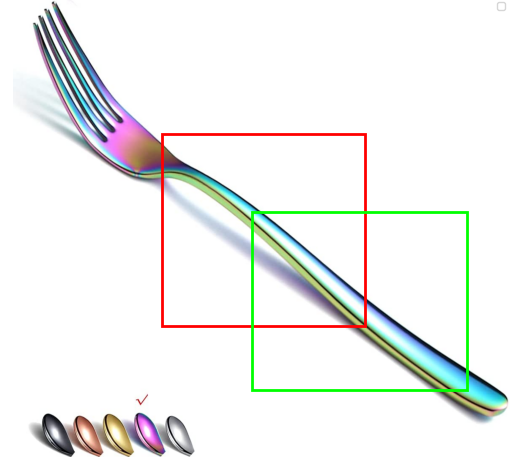} &
        \includegraphics[width=0.32\textwidth, height=4cm]{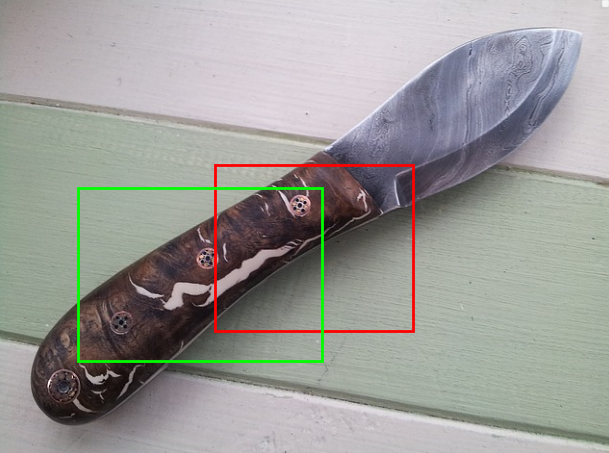} &
        \includegraphics[width=0.32\textwidth, height=4cm]{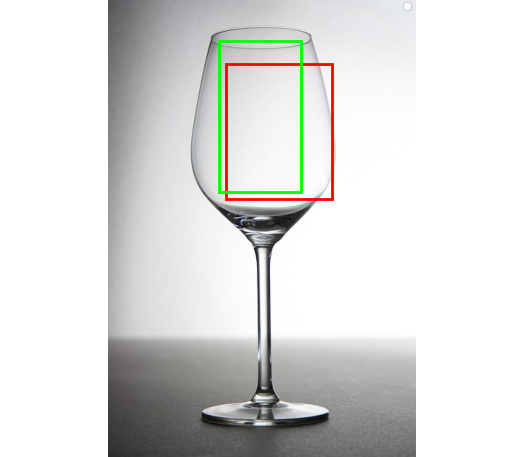} \\ [3pt]
        {\small Hold fork} & {\small Hold knife} & {\small Wash glass} \\[7pt]
        \midrule \\[7pt]
        \includegraphics[width=0.32\textwidth, height=4cm]{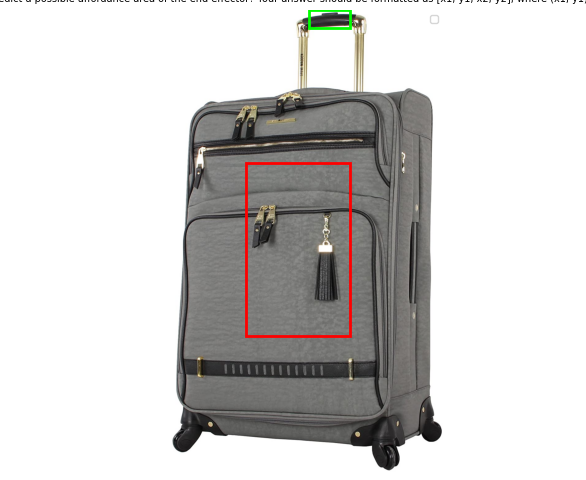} &
        \includegraphics[width=0.32\textwidth, height=4cm]{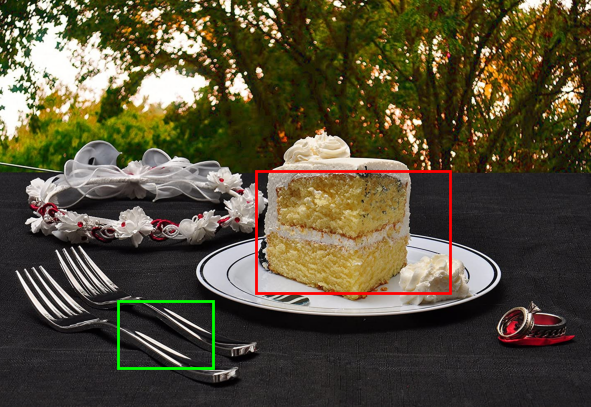} &
        \includegraphics[width=0.32\textwidth, height=4cm]{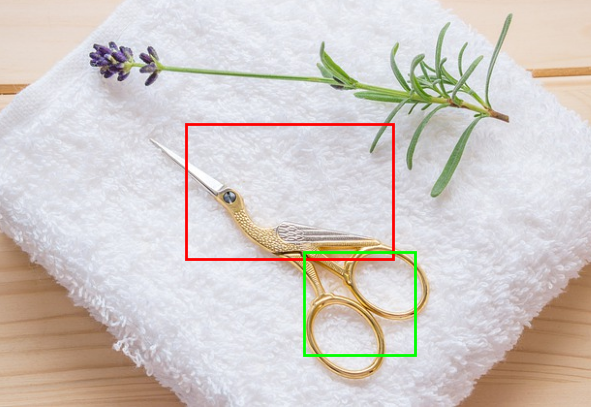} \\ [3pt]
        {\small Drag bag} & {\small Hold fork} & {\small Hold scissors} \\
    \end{tabular}
    \caption{\textbf{Additional visualizations of diverse affordance regions  predicted by LLaVA-OV-0.5B + Geo-Head.} Green box represents the ground-truth and the red box represents the prediction. The visualizations in the first three rows demonstrate that our geometry head effectively generates affordance boxes based on the observation and task instruction. The fourth row presents some failure cases.}
    \label{fig:llava_aff}
\end{figure*}

Figures~\ref{fig:qwen_traj}--\ref{fig:llava_aff} show additional trajectory and affordance predictions for both Qwen2-VL-2B and LLaVA-OV-0.5B across a diverse set of manipulation scenarios.
\label{sec:supp_qual}

\end{document}